\documentclass{article}
\usepackage{epsfig}
\usepackage{multirow}
\usepackage{amssymb}

\renewenvironment{abstract}{\section*{Abstract}\small}{}
\newtheorem{definition}{Definition}

\newtheorem{lemma}[definition]{Lemma}
\newtheorem{notation}[definition]{Notation}

\newtheorem{proposition}[definition]{Proposition}

\newtheorem{remark}[definition]{Remark}

\makeatletter
\renewcommand{\@begintheorem}[2]{ 
\trivlist\item[\hskip\labelsep{\bf #1\ #2}]}
\renewcommand{\@opargbegintheorem}[3]{\trivlist
\item[\hskip \labelsep{\bf #1\ #2\ (#3)}]}
\makeatother
\newtheorem{proof}{Proof}

\newcommand{\qed}{\nobreak \ifvmode \relax \else
\ifdim\lastskip<1.5em \hskip-\lastskip
\hskip1.5em plus0em minus0.5em \fi \nobreak
\vrule height0.75em width0.5em depth0.25em\fi}

\newcommand{\M}[1]{M_{#1}}
\newcommand{\MM}[2]{M_{#1, #2}}
\newcommand{\MMM}[3]{M_{#1, #2, #3}}
\newcommand{\Mp}[1]{M'_{#1}}
\newcommand{\Mi}[2]{M^{#1}_{#2}}
\newcommand{\n}[1]{n(#1)}
\newcommand{\bet}[1]{\beta_{#1}}
\newcommand{\beti}[2]{\beta^{#1}_{#2}}
\newcommand{\F}[1]{F(#1)}
\newcommand{\monH}[1]{H(#1)}
\newcommand{\monHi}[2]{H_{#1}(#2)}
\newcommand{\G}[1]{G(#1)}
\newcommand{\T}[1]{T(#1)}
\newcommand{\TT}[2]{T(#1, #2)}
\newcommand{\Td}[1]{T_d(#1)}
\newcommand{\Tc}[1]{T_c(#1)}
\newcommand{\C}[2]{{#1}(#2)}
\newcommand{\CC}[3]{{#1}(#2, #3)}
\newcommand{\CCC}[4]{{#1}(#2, #3, #4)}
\newcommand{\CCCC}[5]{{#1}(#2, #3, #4, #5)}
\newcommand{\mup}[1]{\mu_{#1}}
\newcommand{\mupp}[2]{\mu_{#1}(#2)}
      
\author{Jonathan Ben-Naim
\\
LIF, CNRS
\\
CMI 39, rue Joliot-Curie
\\
F-13453 Marseille Cedex 13, France
\\
jbennaim@lif.univ-mrs.fr}
\title{Preferential and Preferential-discriminative Consequence Relations
\footnote{
This is an updated version of the paper of the same title published in {\it The Journal of Logic and Computation}.
This version just contains a better presentation (so the numbering of definitions and propositions is different).}}
\date{}

\begin{document}
\maketitle

\begin{abstract}
The present paper investigates
consequence relations that are both non-monotonic and paraconsistent.
More precisely, we put the focus on preferential consequence relations,
i.e. those relations that can be defined by a binary preference relation on states labelled by valuations.
We worked with a general notion of valuation that covers e.g. the classical valuations
as well as certain kinds of many-valued valuations.
In the many-valued cases, preferential consequence relations are paraconsistant (in addition to be non-monotonic),
i.e. they are capable of drawing reasonable conclusions which contain contradictions.
The first purpose of this paper is to provide in our general framework
syntactic characterizations of several families of preferential relations.
The second and main purpose is to provide, again in our general framework,
characterizations of several families of preferential-discriminative consequence relations.
They are defined exactly as the plain version, but any conclusion such that its negation is also a conclusion is rejected
(these relations bring something new essentially in the many-valued cases).
\end{abstract}

\newpage
\section{Introduction} \label{PREFintro}

In many situations, we are confronted with incomplete and/or inconsistent information and the
classical consequence relation proves to be insufficient.
Indeed, in case of inconsistent information, it leads to accept every formula
as a conclusion, which amounts to loose the whole information.
Therefore, we need other relations leading
to non-trivial conclusions in spite of the presence of contradictions.
So, several paraconsistent consequence relations have been developed.
In the present paper, we will pay attention in particular to certain
many-valued ones
\cite{Belnap1, Belnap2, DottavianoDaCosta1, CarnielliMarcosAmo1, CarnielliMarcosAmo2, ArieliAvron2, ArieliAvron3, ArieliAvron1}. They are defined in frameworks where valuations
can assign more than two different truth values to formulas.
In fact, they tolerate contradictions within the conclusions, but reject
the principle of explosion according to which a single contradiction entails the deduction
of every formula.

In case of incomplete information, the classical consequence relation
also shows its limits. Indeed, no risk is taken, the conclusions are sure, but too few.
We need other relations leading to accept as conclusions formulas that are not necessarily sure,
but still plausible. Eventually, some ``hasty'' conclusions will be rejected later, in the presence of additional information.
So, a lot of plausible (generally non-monotonic) consequence relations have been developed.
{\it Choice functions} are central tools to define plausible relations 
\cite{Chernoff1, Arrow1, Sen1, AizermanMalishevski1, Lehmann1, Lehmann2, Schlechta2, Schlechta5}.
Indeed, suppose we have at our disposal a function $\mu$, called a choice function, which chooses
in any set of valuations $V$, those elements which are preferred,
not necessarily in the absolute sense, but when the valuations in $V$ are the only ones
under consideration. Then, we can define a plausible consequence relation in the following natural way:
a formula $\alpha$ follows from a set of formulas $\Gamma$ iff
every model for $\Gamma$ chosen by $\mu$ is a model for $\alpha$.

In the present paper, we put the focus on a particular family of choice functions.
Let us present it. Suppose we are given a binary preference relation $\prec$ on states labelled by valuations
(in the style of e.g. \cite{KrausLehmannMagidor1, Schlechta5}).
This defines naturally a choice function.
Indeed, choose in any set of valuations $V$, each element that labels a state which is $\prec$-preferred
among those states which are labelled by the elements of $V$.
Those choice functions which can be defined in this manner constitute the aforementioned family.
The consequence relations defined by this family will be called {\it preferential consequence relations}.

For a long time, research efforts on paraconsistent relations and plausible relations
were separated. However, in many applications, the information is both incomplete and
inconsistent. For instance, the semantic web or big databases inevitably contain inconsistencies.
This can be due to human or material imperfections as well as contradictory sources of information.
On the other hand, neither the web nor big databases can contain ``all'' information.
Indeed, there are rules of which the exceptions cannot be enumerated.
Also, some information might be left voluntarily vague or in concise form.
Consequently, consequence relations that are both paraconsistent and plausible are useful
to reason in such applications.

Such relations first appear in e.g. \cite{Priest1, Batens1, KiferLozinskii1, ArieliAvron4, MarquisKonieczny1}.
The idea begins by taking a many-valued framework to get paraconsistency.
Then, only those models that are most preferred according to
some particular binary preference relation on valuations (in the style of \cite{Shoham1, Shoham2})
are relevant for making inference, which provides plausibility.
In \cite{AvronLev2, AvronLev3},
A. Avron and I. Lev generalized the study to families of binary preference relations which compare two valuations
using, for each of them, this part of a certain set of formulas it satisfies.
The present paper follows this line of research by combining many-valued frameworks
and choice functions.

More explicitly, we will investigate preferential consequence relations in a general framework.
According to the different assumptions which will be made about the latter,
it will cover various kinds of frameworks, including e.g. the classical propositional one
as well as certain many-valued ones. Moreover, in the many-valued frameworks, preferential relations
are paraconsistent (in addition to be plausible).
However, they do not satisfy the Disjunctive Syllogism (from $\alpha$ and $\neg\alpha \vee \beta$ we can conclude $\beta$),
whilst they satisfy it in classical framework.

In addition, we will investigate {\it preferential-discriminative consequence relations}.
They are defined exactly as the plain version, but any conclusion such that its negation is also a conclusion is rejected.
In the classical framework, they do not bring something really new.
Indeed, instead of concluding everything in the face of inconsistent information,
we will simply conclude nothing.
On the other hand, in the many-valued frameworks, where
the conclusions are non-trivial even from inconsistent information,
the discriminative version will reject the contradictions among them, rendering them
all the more rational.

The contribution of the present paper can now be summarized in one sentence:
we characterized, in a general framework, several (sub)families of preferential(-discriminative)
consequence relations.
In many cases, our characterizations are purely syntactic.
This has a lot of advantages, let us quote some important ones.
Take some syntactic conditions that characterize a family of those consequence relations.
This gives a syntactic point of view on this family defined semantically,
which enables us to compare it to conditions known on the ``market'',
and thus to other consequence relations.
This can also give rise to questions like:
if we modified the conditions in such and such a natural-looking way,
what would happen on the semantic side?
More generally, this can open the door to questions that would not easily come
to mind otherwise or to techniques of proof that could not have been employed in the
semantic approach.

Several characterizations can be found in the literature
for preferential relations
(e.g. \cite{Gabbay1, Makinson5, Makinson4, KrausLehmannMagidor1, LehmannMagidor1, Lehmann1, Lehmann2, Schlechta2, Schlechta3, Schlechta1, Schlechta5}).
We will provide some new ones, though to do so we have been
strongly inspired by techniques of K. Schlechta \cite{Schlechta5}.
In fact, our innovation is rather related to the discriminative version.
To the author knowledge, the present paper is the first systematic work of characterization
for preferential-discriminative consequence relations.

The rest of the paper is organized as follows.
In Section~\ref{PREFframework}, we introduce our general framework
and the different assumptions which sometimes will be made about it.
We will see that it covers in particular the many-valued frameworks of the well-known paraconsistent logics
$\cal FOUR$ and $J_3$.
In Section~\ref{PREFchoicefun}, we present choice functions and some of their well-known properties.
In Section~\ref{PREFintroprefCR}, we define preferential(-discriminative) consequence
relations and give examples in both the classical and the many-valued frameworks.
We will also recall a characterization which involves the well-known system $\bf P$ of
Kraus, Lehmann, and Magidor.
In section~\ref{PREFcontribu}, we provide our characterizations.
Finally, we conclude in Section~\ref{PREFconclu}.

\section{Background} \label{PREFprelim}

\subsection{Semantic structures} \label{PREFframework}

\subsubsection{Definitions and properties}

We will work with general formulas, valuations, and satisfaction.
A similar approach has been taken in two well-known papers \cite{Makinson2, Lehmann2}.

\begin{definition}
We say that $\cal S$ is a {\it semantic structure} iff
$\cal S = \langle {\cal F}, {\cal V}, \models \rangle$ where
$\cal F$ is a set,
$\cal V$ is a set,
and $\models$ is a relation on ${\cal V} \times {\cal F}$.
\end{definition}
Intuitively, $\cal F$ is a set of formulas,
$\cal V$ a set of valuations for these formulas, and
$\models$ a satisfaction relation for these objects (i.e. $v \models \alpha$ means the formula $\alpha$ is satisfied in the valuation $v$,
i.e. $v$ is a model for $\alpha$).

\begin{notation}
Let $\langle {\cal F}, {\cal V}, \models \rangle$ be a semantic structure,
$\Gamma \subseteq {\cal F}$, and $V \subseteq {\cal V}$. Then,
\\
$\M{\Gamma} := \lbrace v \in {\cal V} : \forall \: \alpha \in \Gamma$, $v \models \alpha \rbrace$,
\\
$\T{V} := \lbrace \alpha \in {\cal F} : V \subseteq \M{\alpha} \rbrace$,
\\
${\bf D} := \lbrace V \subseteq {\cal V} : \exists \: \Gamma \subseteq {\cal F}, \M{\Gamma} = V \rbrace$.
\\
Suppose $\cal L$ is a language, $\neg$ a unary connective of $\cal L$, and 
$\cal F$ the set of all wffs of $\cal L$. Then,
\\
$\Td{V} := \lbrace \alpha \in {\cal F} : V \subseteq \M{\alpha}$ and $V \not \subseteq \M{\neg\alpha} \rbrace$,
\\
$\Tc{V} := \lbrace \alpha \in {\cal F} : V \subseteq \M{\alpha}$ and $V \subseteq \M{\neg\alpha} \rbrace$,
\\
${\bf C} := \lbrace V \subseteq {\cal V} : \forall \: \alpha \in {\cal F}$, $V \not\subseteq \M{\alpha}$ or
$V \not\subseteq \M{\neg\alpha} \rbrace$.
\end{notation}
Intuitively, $\M{\Gamma}$ is the set of all models for $\Gamma$
and $\T{V}$ the set of all formulas satisfied in $V$.
Every element of $\T{V}$ belongs either to $\Td{V}$ or $\Tc{V}$,
according to whether its negation is also in $\T{V}$.
$\bf D$ is the set of all those sets of valuations that are definable by a set of formulas
and $\bf C$ the set of all those sets of valuations that do not satisfy both a formula and its negation.
As usual, $\MM{\Gamma}{\alpha}$, $\TT{V}{v}$ stand for respectively
$\M{\Gamma \cup \lbrace \alpha \rbrace}$, $\T{V \cup \lbrace v \rbrace}$, etc.

\begin{remark}
The notations $\M{\Gamma}$, $\T{V}$, etc. should contain the semantic structure on which
they are based. To increase readability, we will omit it. There will never be any ambiguity.
We will omit similar things with other notations in the sequel, for the same reason.
\end{remark}
A semantic structure defines a basic consequence relation:

\begin{notation}
We denote by $\cal P$ the power set operator.
\\
Let $\langle {\cal F}, {\cal V}, \models \rangle$ be a semantic structure.
\\
We denote by $\vdash$ the relation on ${\cal P}({\cal F}) \times {\cal F}$
such that $\forall \: \Gamma \subseteq {\cal F}$, $\forall \: \alpha \in {\cal F}$,
$$\Gamma \vdash \alpha\; \textrm{iff}\; \M{\Gamma} \subseteq \M{\alpha}.$$
Let $\mid\!\sim$ be a relation on ${\cal P}({\cal F}) \times {\cal F}$. Then,
\\
$\C{\mid\!\sim}{\Gamma} := \lbrace \alpha \in {\cal F} : \Gamma \mid\!\sim \alpha \rbrace$.
\\
Suppose $\cal L$ is a language, $\neg$ a unary connective of $\cal L$,
$\cal F$ the set of all wffs of $\cal L$, and $\Gamma \subseteq {\cal F}$.
\\
Then, we say that $\Gamma$ is {\it consistent} iff $\forall \: \alpha \in {\cal F}$, $\Gamma \not\vdash \alpha$ or $\Gamma \not\vdash \neg\alpha$.
\end{notation}
The following trivial facts hold, we will use them implicitly in the sequel:
\begin{remark}
Let $\langle {\cal F}, {\cal V}, \models \rangle$ be a semantic structure and $\Gamma, \Delta \subseteq {\cal F}$. Then:
\\
$\M{\Gamma, \Delta} = \M{\Gamma} \cap \M{\Delta}$;
\\
$\C{\vdash}{\Gamma} = \T{\M{\Gamma}}$;
\\
$\M{\Gamma} = \M{\C{\vdash}{\Gamma}}$;
\\
$\Gamma \subseteq \C{\vdash}{\Delta}$ iff $\C{\vdash}{\Gamma} \subseteq \C{\vdash}{\Delta}$
iff $\M{\Delta} \subseteq \M{\Gamma}$.
\end{remark}
Sometimes, we will need some of the following assumptions about a semantic structure:

\begin{definition}
Suppose $\langle {\cal F}, {\cal V}, \models \rangle$ is a semantic structure.
\\
Then, define the following assumptions about it:
\begin{description}
\item[$(A1)$] $\cal V$ is finite.
\end{description}
Suppose $\cal L$ is a language, $\neg$ a unary connective of $\cal L$,
and $\cal F$ the set of all wffs of $\cal L$. Then, define:
\begin{description}
\item[$(A2)$] $\forall \: \Gamma \subseteq {\cal F}$, $\forall \: \alpha \in {\cal F}$,
if $\alpha \not\in \T{\M{\Gamma}}$ and $\neg\alpha \not\in \T{\M{\Gamma}}$, then $\M{\Gamma} \cap \M{\alpha} \not\subseteq \M{\neg\alpha}$.
\end{description}
Suppose $\vee$ and $\wedge$ are binary connectives of $\cal L$. Then, define:
\begin{description}
\item[$(A3)$] $\forall \: \alpha, \beta \in {\cal F}$, we have:
\\
$\M{\alpha \vee \beta} = \M{\alpha} \cup \M{\beta}$;
\\
$\M{\alpha \wedge \beta} = \M{\alpha} \cap \M{\beta}$;
\\
$\M{\neg\neg\alpha} = \M{\alpha}$;
\\
$\M{\neg(\alpha \vee \beta)} = \M{\neg\alpha \wedge \neg\beta}$;
\\
$\M{\neg(\alpha \wedge \beta)} = \M{\neg\alpha \vee \neg\beta}$.
\end{description}
\end{definition}
Clearly, those assumptions are satisfied by classical semantic structures, i.e.
structures where $\cal F$, $\cal V$, and $\models$ are classical.
In addition, we will see, in Sections~\ref{PREFfourframework} and \ref{PREFj3framework},
that they are satisfied also by certain many-valued semantic structures.

\subsubsection{The semantic structure defined by $\cal FOUR$} \label{PREFfourframework}

The logic $\cal FOUR$ was introduced by N. Belnap in \cite{Belnap2, Belnap1}.
This logic is useful to deal with inconsistent information.
Several presentations are possible, depending on the language under consideration.
For the needs of the present paper, a classical propositional language will be sufficient.
The logic has been investigated intensively in e.g. \cite{ArieliAvron2, ArieliAvron3, ArieliAvron1},
where richer languages, containing an implication connective $\supset$ (first introduced by A. Avron \cite{Avron1}),
were considered.

\begin{notation}
We denote by $\cal A$ a set of propositional symbols (or atoms).
\hfill \\
We denote by ${\cal L}_c$ the classical propositional language containing $\cal A$,
the usual constants $false$ and $true$, and the usual connectives $\neg$, $\vee$, and $\wedge$.
\hfill \\
We denote by ${\cal F}_c$ the set of all wffs of ${\cal L}_c$.
\end{notation}
We recall a possible meaning for the logic $\cal FOUR$
(more details can be found in  \cite{CarnielliLimaMarques1, Belnap2, Belnap1}).
Consider a system in which there are, on the one hand, sources of information
and, on the other hand, a processor that listens to them.
The sources provide information about the atoms only,
not about the compound formulas.
For each atom $p$, there are exactly four possibilities:
either the processor is informed (by the sources, taken as a whole) that $p$ is true; or he is informed that $p$ is false;
or he is informed of both; or he has no information about $p$.

\begin{notation}
Denote by $0$ and $1$ the classical truth values and define:
\hfill \\
${\bf f} := \lbrace 0 \rbrace$;\qquad ${\bf t} := \lbrace 1 \rbrace$;\qquad
$\top := \lbrace 0, 1 \rbrace$;\qquad $\bot := \emptyset$.
\end{notation}
The global information given by the sources to the processor can be modelled by
a function $s$ from ${\cal A}$ to $\lbrace {\bf f}, {\bf t}, \top, \bot  \rbrace$.
Intuitively, $1 \in s(p)$ means the processor is informed that $p$ is true,
whilst $0 \in s(p)$ means he is informed that $p$ is false.

Then, the processor naturally builds information about the compound formulas from $s$.
Before he starts to do so, the situation can be
be modelled by a function $v$ from ${\cal F}_c$ to $\lbrace {\bf f}, {\bf t}, \top, \bot  \rbrace$
which agrees with $s$ about the atoms
and which assigns $\bot$ to all compound formulas.
Now, take $p$ and $q$ in $\cal A$ and suppose $1 \in v(p)$ or $1 \in v(q)$.
Then, the processor naturally adds $1$ to $v(p\vee q)$.
Similarly, if $0 \in v(p)$ and $0 \in v(q)$, then he adds $0$ in $v(p \vee q)$.
Of course, such rules hold for $\neg$ and $\wedge$ too.

Suppose all those rules are applied recursively to all compound formulas.
Then, $v$ represents the ``full'' (or developed) information given by the sources to the processor.
Now, the valuations of the logic $\cal FOUR$ can be defined as exactly those functions
that can be built
in this manner (i.e. like $v$) from some information sources. More formally,

\begin{definition}
We say that $v$ is a {\it four-valued valuation}
iff $v$ is a function from ${\cal F}_c$ to $\lbrace {\bf f}, {\bf t}, \top, \bot  \rbrace$
such that $v(true) = {\bf t}$, $v(false) = {\bf f}$ and $\forall \: \alpha, \beta \in {\cal F}_c$,
\hfill \\
$1 \in v(\neg\alpha)$ iff $0 \in v(\alpha)$;
\hfill \\
$0 \in v(\neg\alpha)$ iff $1 \in v(\alpha)$;
\hfill \\
$1 \in v(\alpha \vee \beta)$ iff $1 \in v(\alpha)$ or $1 \in v(\beta)$;
\hfill \\
$0 \in v(\alpha \vee \beta)$ iff $0 \in v(\alpha)$ and $0 \in v(\beta)$;
\hfill \\
$1 \in v(\alpha \wedge \beta)$ iff $1 \in v(\alpha)$ and $1 \in v(\beta)$;
\hfill \\
$0 \in v(\alpha \wedge \beta)$ iff $0 \in v(\alpha)$ or $0 \in v(\beta)$.
\hfill \\
We denote by ${\cal V}_4$ the set of all four-valued valuations.
\end{definition}
The definition may become more accessible if we see the four-valued valuations
as those functions that satisfy Tables 1, 2, and 3 below:
\begin{center}
\begin{tabular}{cc}
$v(\alpha)$ & $v(\neg\alpha)$\\
\hline
\multicolumn{1}{|c|}{$\bf f$} & \multicolumn{1}{|c|}{$\bf t$}\\
\multicolumn{1}{|c|}{$\bf t$} & \multicolumn{1}{|c|}{$\bf f$} \\
\multicolumn{1}{|c|}{$\top$} & \multicolumn{1}{|c|}{$\top$} \\
\multicolumn{1}{|c|}{$\bot$} & \multicolumn{1}{|c|}{$\bot$} \\
\hline
\multicolumn{2}{c}{Table 1.}
\end{tabular}
\begin{tabular}{cccccc}
 & & \multicolumn{4}{c}{$v(\beta)$}\\
 \cline{3-6}
 & & \multicolumn{1}{|c}{$\bf f$} & \multicolumn{1}{c}{$\bf t$} & \multicolumn{1}{c}{$\top$} & \multicolumn{1}{c|}{$\bot$}  \\
\cline{2-6}
 \multirow{4}{*}{$v(\alpha)$} &  \multicolumn{1}{|c|}{$\bf f$}  & \multicolumn{1}{|c}{$\bf f$}    & \multicolumn{1}{c}{$\bf t$}
 & \multicolumn{1}{c}{$\top$} & \multicolumn{1}{c|}{$\bot$}        \\
 & \multicolumn{1}{|c|}{$\bf t$}  & \multicolumn{1}{|c}{$\bf t$}    & \multicolumn{1}{c}{$\bf t$} 
 & \multicolumn{1}{c}{$\bf t$} & \multicolumn{1}{c|}{$\bf t$}        \\
& \multicolumn{1}{|c|}{$\top$} & \multicolumn{1}{|c}{$\top$} & \multicolumn{1}{c}{$\bf t$}
& \multicolumn{1}{c}{$\top$} & \multicolumn{1}{c|}{$\bf t$}      \\
& \multicolumn{1}{|c|}{$\bot$} & \multicolumn{1}{|c}{$\bot$} & \multicolumn{1}{c}{$\bf t$}
& \multicolumn{1}{c}{$\bf t$} & \multicolumn{1}{c|}{$\bot$}       \\
\cline{2-6}
&  & \multicolumn{4}{c}{$v(\alpha \vee \beta)$}\\
\multicolumn{6}{c}{Table 2.}
\end{tabular}
\begin{tabular}{cccccc}
 & & \multicolumn{4}{c}{$v(\beta)$}\\
 \cline{3-6}
 & & \multicolumn{1}{|c}{$\bf f$} & \multicolumn{1}{c}{$\bf t$} & \multicolumn{1}{c}{$\top$} & \multicolumn{1}{c|}{$\bot$}  \\
\cline{2-6}
 \multirow{4}{*}{$v(\alpha)$} &  \multicolumn{1}{|c|}{$\bf f$}  & \multicolumn{1}{|c}{$\bf f$}    & \multicolumn{1}{c}{$\bf f$} 
  & \multicolumn{1}{c}{$\bf f$} & \multicolumn{1}{c|}{$\bf f$}        \\
 & \multicolumn{1}{|c|}{$\bf t$}  & \multicolumn{1}{|c}{$\bf f$}    & \multicolumn{1}{c}{$\bf t$}    
 & \multicolumn{1}{c}{$\top$} & \multicolumn{1}{c|}{$\bot$}        \\
& \multicolumn{1}{|c|}{$\top$} & \multicolumn{1}{|c}{$\bf f$} & \multicolumn{1}{c}{$\top$}
& \multicolumn{1}{c}{$\top$} & \multicolumn{1}{c|}{$\bf f$}      \\
& \multicolumn{1}{|c|}{$\bot$} & \multicolumn{1}{|c}{$\bf f$} & \multicolumn{1}{c}{$\bot$}
 & \multicolumn{1}{c}{$\bf f$} & \multicolumn{1}{c|}{$\bot$}       \\
\cline{2-6}
& & \multicolumn{4}{c}{$v(\alpha \wedge \beta)$}\\
\multicolumn{6}{c}{Table 3.}
\end{tabular}
\end{center}
In the logic $\cal FOUR$, a formula $\alpha$ is considered to be satisfied iff the processor is informed
that it is true (it does not matter whether he is also informed that $\alpha$ is false).

\begin{notation}
We denote by $\models_4$ the relation on ${\cal V}_4 \times {\cal F}_c$ such that
$\forall \: v \in {\cal V}_4$, $\forall \: \alpha \in {\cal F}_c$, we have
\hfill \\
$v \models_4 \alpha$ iff $1 \in v(\alpha)$.
\end{notation}
Proof systems for the consequence relation $\vdash$
based on the semantic structure $\langle {\cal F}_c, {\cal V}_4, \models_4 \rangle$
(i.e. the semantic structure defined by $\cal FOUR$)
can be found in e.g. \cite{ArieliAvron2, ArieliAvron3, ArieliAvron1}.

Note that the $\cal FOUR$ semantic structure satisfies $(A3)$.
In addition, if $\cal A$ is finite, then $(A1)$ is also satisfied.
However, $(A2)$ is not satisfied by this structure.
In Section~\ref{PREFj3framework}, we turn to a many-valued semantic structure which satisfies $(A2)$.

\subsubsection{The semantic structure defined by $J_3$} \label{PREFj3framework}

The logic $J_3$ was introduced in \cite{DottavianoDaCosta1} to answer a question posed in 1948 by S. Ja\'{s}kowski, who was interested in systematizing theories capable of containing contradictions, especially if they occur in dialectical reasoning. The step from informal reasoning under contradictions and formal reasoning with databases and information was done in \cite{CarnielliMarcosAmo1} (also specialized for real database models in \cite{CarnielliMarcosAmo2}), where another formulation of $J_3$ called {\bf LFI1} was introduced, and its first-order version, semantics and proof theory were studied in detail.
Investigations of $J_3$ have also been made in e.g. \cite{Avron1},
where richer languages than our ${\cal L}_c$ were considered.

The valuations of the logic $J_3$ can be given the same meaning as those of the logic $\cal FOUR$,
except that the consideration is restricted to those sources which
always give some information about an atom.
More formally,

\begin{definition}
We say that $v$ is a {\it three-valued valuation} iff $v$ is a function from ${\cal F}_c$ to $\lbrace {\bf f}, {\bf t}, \top \rbrace$
such that $v(true) = {\bf t}$, $v(false) = {\bf f}$ and $\forall \: \alpha, \beta \in {\cal F}_c$,
\hfill \\
$1 \in v(\neg\alpha)$ iff $0 \in v(\alpha)$;
\hfill \\
$0 \in v(\neg\alpha)$ iff $1 \in v(\alpha)$;
\hfill \\
$1 \in v(\alpha \vee \beta)$ iff $1 \in v(\alpha)$ or $1 \in v(\beta)$;
\hfill \\
$0 \in v(\alpha \vee \beta)$ iff $0 \in v(\alpha)$ and $0 \in v(\beta)$;
\hfill \\
$1 \in v(\alpha \wedge \beta)$ iff $1 \in v(\alpha)$ and $1 \in v(\beta)$;
\hfill \\
$0 \in v(\alpha \wedge \beta)$ iff $0 \in v(\alpha)$ or $0 \in v(\beta)$.
\hfill \\
We denote by ${\cal V}_3$ the set of all three-valued valuations.
\end{definition}
As previously, the definition may become more accessible
if we see the three-valued valuations as those functions that satisfy Tables 4, 5, and 6 below:

\begin{center}
\begin{tabular}{cc}
$v(\alpha)$ & $v(\neg\alpha)$\\
\hline
\multicolumn{1}{|c|}{$\bf f$} & \multicolumn{1}{|c|}{$\bf t$}\\
\multicolumn{1}{|c|}{$\bf t$} & \multicolumn{1}{|c|}{$\bf f$} \\
\multicolumn{1}{|c|}{$\top$} & \multicolumn{1}{|c|}{$\top$} \\
\hline
\multicolumn{2}{c}{Table 4.}
\end{tabular}
\begin{tabular}{ccccc}
 & & \multicolumn{3}{c}{$v(\beta)$}\\
 \cline{3-5}
 & & \multicolumn{1}{|c}{$\bf f$} & \multicolumn{1}{c}{$\bf t$} & \multicolumn{1}{c|}{$\top$} \\
\cline{2-5}
 \multirow{3}{*}{$v(\alpha)$} &  \multicolumn{1}{|c|}{$\bf f$}  & \multicolumn{1}{|c}{$\bf f$}    & \multicolumn{1}{c}{$\bf t$}    & \multicolumn{1}{c|}{$\top$}    \\
 & \multicolumn{1}{|c|}{$\bf t$}  & \multicolumn{1}{|c}{$\bf t$}    & \multicolumn{1}{c}{$\bf t$}    & \multicolumn{1}{c|}{$\bf t$}    \\
& \multicolumn{1}{|c|}{$\top$} & \multicolumn{1}{|c}{$\top$} & \multicolumn{1}{c}{$\bf t$} & \multicolumn{1}{c|}{$\top$}  \\
\cline{2-5}
&  & \multicolumn{3}{c}{$v(\alpha \vee \beta)$}\\
\multicolumn{5}{c}{Table 5.}
\end{tabular}
\begin{tabular}{ccccc}
 & & \multicolumn{3}{c}{$v(\beta)$}\\
 \cline{3-5}
 & & \multicolumn{1}{|c}{$\bf f$} & \multicolumn{1}{c}{$\bf t$} & \multicolumn{1}{c|}{$\top$} \\
\cline{2-5}
 \multirow{3}{*}{$v(\alpha)$} &  \multicolumn{1}{|c|}{$\bf f$}  & \multicolumn{1}{|c}{$\bf f$}   & \multicolumn{1}{c}{$\bf f$}    & \multicolumn{1}{c|}{$\bf f$}    \\
 & \multicolumn{1}{|c|}{$\bf t$}  & \multicolumn{1}{|c}{$\bf f$}    & \multicolumn{1}{c}{$\bf t$}    & \multicolumn{1}{c|}{$\top$}    \\
& \multicolumn{1}{|c|}{$\top$} & \multicolumn{1}{|c}{$\bf f$} & \multicolumn{1}{c}{$\top$} & \multicolumn{1}{c|}{$\top$}  \\
\cline{2-5}
& & \multicolumn{3}{c}{$v(\alpha \wedge \beta)$}\\
\multicolumn{5}{c}{Table 6.}
\end{tabular}
\end{center}
We turn to the satisfaction relation.

\begin{notation}
We denote by $\models_3$ the relation on ${\cal V}_3 \times {\cal F}_c$ such that
$\forall \: v \in {\cal V}_3$, $\forall \: \alpha \in {\cal F}_c$, we have
\hfill \\
$v \models_3 \alpha$ iff $1 \in v(\alpha)$.
\end{notation}
Proof systems for the consequence relation $\vdash$
based on the semantic structure $\langle {\cal F}_c, {\cal V}_3, \models_3 \rangle$
(i.e. the semantic structure defined by $J_3$) have been provided in e.g.
\cite{Avron1, DottavianoDaCosta1} and chapter IX of \cite{Epstein1}.
The $J_3$ structure satisfies $(A3)$ and $(A2)$.
In addition, if $\cal A$ is finite, then it satisfies $(A1)$ too.

\subsection{Choice functions} \label{PREFchoicefun}

\subsubsection{Definitions and properties}

In many situations, an agent has some way to choose
in any set of valuations $V$, those elements that are preferred (the bests, the more normal, etc.),
not necessarily in the absolute sense, but when the valuations in $V$ are the only ones under consideration.
In Social Choice, this is modelled by choice functions
\cite{Chernoff1, Arrow1, Sen1, AizermanMalishevski1, Lehmann1, Lehmann2}.

\begin{definition}
Let $\cal V$ be a set, ${\bf V} \subseteq {\cal P}({\cal V})$, ${\bf W} \subseteq {\cal P}({\cal V})$, and
$\mu$ a function from $\bf V$ to $\bf W$.
\\
We say that $\mu$ is a {\it choice function} iff $\forall \: V \in {\bf V}$, $\mu(V) \subseteq V$.
\end{definition}
Several properties for choice functions have been put in evidence by
researchers in Social Choice.
Let us present two important ones (a better presentation can be found in \cite{Lehmann2}).
Suppose $W$ is a set of valuations, $V$ is a subset of $W$,
and $v \in V$ is a preferred valuation of $W$.
Then, a natural requirement is that $v$ is a preferred valuation of $V$.
Indeed, in many situations, the larger a set is, the harder it is to be a preferred element of it,
and he who can do the most can do the least.
This property appears in \cite{Chernoff1} and has been given the name Coherence
in \cite{Moulin1}.

We turn to the second property.
Suppose $W$ is a set of valuations, $V$ is a subset of $W$, and
suppose all the preferred valuations of $W$ belong to $V$.
Then, they are expected to include all the preferred valuations of $V$.
The importance of this property has been put in evidence by \cite{Aizerman1, AizermanMalishevski1}
and has been given the name Local Monotonicity in e.g. \cite{Lehmann2}.

\begin{definition}
Let $\cal V$ be a set, ${\bf V} \subseteq {\cal P}({\cal V})$, ${\bf W} \subseteq {\cal P}({\cal V})$, and
$\mu$ a choice function from $\bf V$ to $\bf W$.
\\
We say that $\mu$ is {\it coherent} iff $\forall \: V, W \in {\bf V}$,
$$\textrm{if}\; V \subseteq W,\; \textrm{then}\; \mu(W) \cap V \subseteq \mu(V).$$
We say that $\mu$ is {\it locally monotonic} (LM) iff $\forall \: V, W \in {\bf V}$,
$$\textrm{if}\; \mu(W) \subseteq V \subseteq W,\; \textrm{then}\; \mu(V) \subseteq \mu(W).$$
\end{definition}
In addition to their intuitive meanings, these properties are important because, as was shown by K.~Schlechta in
\cite{Schlechta1},
they characterize those choice functions that can be defined by a binary preference relation on states
labelled by valuations (in the style of e.g. \cite{KrausLehmannMagidor1}).
We will take a closer look at this in Section~\ref{PREFredefPref}.

When a semantic structure is under consideration, two new properties can be defined.
Each of them conveys a simple and natural meaning.

\begin{definition}
Let $\langle {\cal F}, {\cal V}, \models \rangle$ be a semantic structure,
${\bf V} \subseteq {\cal P}({\cal V})$, ${\bf W} \subseteq {\cal P}({\cal V})$, and
$\mu$ a choice function from $\bf V$ to $\bf W$.
\\
We say that $\mu$ is {\it definability preserving} (DP) iff
$$\forall \: V \in {\bf V} \cap {\bf D},\; \mu(V) \in {\bf D}.$$
Suppose $\cal L$ is a language, $\neg$ a unary connective of $\cal L$,
and $\cal F$ the set of all wffs of $\cal L$.
\\
We say that $\mu$ is {\it coherency preserving} (CP) iff
$$\forall \: V \in {\bf V} \cap {\bf C},\; \mu(V) \in {\bf C}.$$
\end{definition}
Definability Preservation has been put in evidence first in \cite{Schlechta2}.
One of its advantages is that when the choice functions under consideration
satisfy it, we will provide characterizations with purely syntactic conditions.
To the author knowledge, the present paper is the first to introduce Coherency Preservation.
An advantage of this property is that when the choice functions under consideration satisfy it,
we will not need to assume $(A2)$ to show our characterizations (in the discriminative case).

\subsubsection{Preference structures} \label{PREFredefPref}

Binary preference relations on valuations have been investigated by e.g.
B. Hansson to give semantics for deontic logics \cite{Hansson1}.
Y. Shoham rediscovered them to give semantics for plausible non-monotonic logics \cite{Shoham1, Shoham2}.
Then, it seems that Imielinski is one of the first persons to introduce
binary preference relations on states labelled by valuations \cite{Imielinski1}.
They have been used to give more general semantics for plausible non-monotonic logics,
see e.g. \cite{KrausLehmannMagidor1, LehmannMagidor1, Schlechta2, Schlechta3, Schlechta1, Schlechta5}.
Let us present them.

\begin{definition}
We say that $\cal R$ is a preference structure on a set ${\cal V}$ iff ${\cal R} = \langle {\cal S}, l, \prec \rangle$
where $\cal S$ is a set,
$l$ is a function from ${\cal S}$ to $\cal V$, and $\prec$ is a relation on ${\cal S} \times {\cal S}$.
\end{definition}
In fact, preference structures are essentially Kripke structures. The difference lies
in the interpretation of $\prec$. In a Kripke structure, it is seen as an accessibility relation, whilst,
in a preference structure, it is seen as a preference relation.
We recall a possible meaning for preference structures (see
e.g. \cite{KrausLehmannMagidor1, Schlechta5} for details about meaning).
Intuitively, $\cal V$ is a set of valuations for some language $\cal L$ and
$\cal S$ a set of valuations for some language ${\cal L}'$ richer than $\cal L$.
The elements of $\cal S$ are called states.
$l(s)$ corresponds precisely to this part of
$s$ that is about the formulas of $\cal L$ only.
We call $l$ a labelling function.
Finally, $\prec$ is a preference relation, i.e. $s \prec s'$ means
$s$ is preferred to $s'$.

We turn to well-known properties for preference structures.

\begin{definition}
Suppose $\cal V$ is a set, ${\cal R} = \langle {\cal S}, l, \prec \rangle$ is a preference structure on $\cal V$,
$S \subseteq {\cal S}$, $s \in S$, $V \subseteq {\cal V}$, and ${\bf V} \subseteq {\cal P}({\cal V})$.
\\
We say that ${\cal R}$ is {\it transitive} (resp. {\it irreflexive}) iff $\prec$ is transitive (resp. irreflexive).
\\
We say that $s$ is {\it preferred} in $S$ iff $\forall \: s' \in S$, $s' \not\prec s$.
\\
$L(V) := \lbrace s \in {\cal S} : l(s) \in V \rbrace$ (intuitively, $L(V)$ contains the states labelled by the elements of $V$).
\\
We say that $\cal R$ is {\bf V}-{\it smooth} (alias {\bf V}-{\it stoppered}) iff $\forall \: V \in {\bf V}$,
$\forall \: s \in L(V)$,
\\
either $s$ is preferred in $L(V)$ or there exists $s'$ preferred in $L(V)$ such that $s' \prec s$.
\end{definition}
A preference structure defines naturally a choice function.
The idea is to choose in any set of valuations $V$,
each element which labels a state which is preferred among all the states labelled by the elements of $V$.

\begin{definition}
Suppose ${\cal R} = \langle {\cal S}, l, \prec \rangle$ is a preference structure on a set $\cal V$.
\\
We denote by $\mu_{\cal R}$ the function from ${\cal P}({\cal V})$ to ${\cal P}({\cal V})$ such that
$\forall \: V \subseteq {\cal V}$,
$$\mu_{\cal R}(V) = \lbrace v \in V :
\exists \: s \in L(v),\; s\; \textrm{is preferred in}\; L(V) \rbrace.$$
\end{definition}
In~\cite{Schlechta1}, Schlechta showed that Coherence and Local Monotonicity
characterize those choice functions that can be defined by a preference structure.
Details are given in the proposition just below.
It is an immediate corollary of Proposition~2.4, Proposition~2.15, and Fact~1.3 of \cite{Schlechta1}.

\begin{proposition}\label{PREFkarlProp} Taken from \cite{Schlechta1}.
\\
Let $\cal V$ be a set, ${\bf V}$ and ${\bf W}$ subsets of ${\cal P}({\cal V})$,
and $\mu$ a choice function from $\bf V$ to $\bf W$. Then,
\begin{description}
\item[$(0)$] $\mu$ is coherent iff there exists a transitive and irreflexive preference structure
$\cal R$ on $\cal V$ such that $\forall \: V \in {\bf V}$, we have $\mu(V) = \mu_{\cal R}(V)$.
\end{description}
Suppose $\forall \: V, W \in {\bf V}$, we have $V \cup W \in {\bf V}$ and $V \cap W \in {\bf V}$. Then,
\begin{description}
\item[$(1)$] $\mu$ is coherent and LM iff there exists a ${\bf V}$-smooth, transitive, and irreflexive preference structure $\cal R$ on $\cal V$ such that $\forall \: V \in {\bf V}$, we have $\mu(V) = \mu_{\cal R}(V)$.
\end{description}
\end{proposition}
In fact, in \cite{Schlechta1}, the codomain of $\mu$ is required to be its domain: $\bf V$.
However, this plays no role in the proofs.
Therefore, verbatim the same proofs are valid when the codomain of
$\mu$ is an arbitrary subset $\bf W$ of ${\cal P}({\cal V})$.
Both myself and Schlechta checked it.

\subsection{Preferential(-discriminative) consequence relations}\label{PREFintroprefCR}

\subsubsection{Definitions}

Suppose we are given a semantic structure and a choice function $\mu$ on the valuations.
Then, it is natural to conclude a formula $\alpha$ from a set of formulas $\Gamma$ iff
every model for $\Gamma$ chosen by $\mu$ is a model for $\alpha$.
More formally:

\begin{definition}\label{PREFdefprefrel}
Suppose ${\cal S} = \langle {\cal F}, {\cal V}, \models \rangle$ is a semantic structure and
$\mid\!\sim$ a relation on ${\cal P}({\cal F}) \times {\cal F}$.
\\
We say that $\mid\!\sim$ is a {\it preferential consequence relation} iff there exists a coherent choice function $\mu$
from $\bf D$ to ${\cal P}({\cal V})$
such that $\forall \: \Gamma \subseteq {\cal F}$, $\forall \: \alpha \in {\cal F}$,
$$\Gamma \mid\!\sim \alpha\; \textrm{iff}\; \mu(\M{\Gamma}) \subseteq \M{\alpha}.$$
In addition, if $\mu$ is LM, DP, etc., then so is $\mid\!\sim$.
\end{definition}
These consequence relations are called ``preferential'' because, in the light of Proposition~\ref{PREFkarlProp}, they
can be defined equivalently with preference structures, instead of coherent choice functions.
They lead to ``jump'' to plausible conclusions
which will eventually be withdrawn later, in the presence of additional information.
Therefore, they are useful to deal with incomplete information.
We will give an example with a classical semantic structure in Section~\ref{PREFclassicalexample}.

In addition, if a many-valued semantic structure is considered,
they lead to rational and non-trivial
conclusions is spite of the presence of contradictions and are thus useful to treat
both incomplete and inconsistent information. However, they will not satisfy
the Disjunctive Syllogism.
We will give an example with the $\cal FOUR$ semantic structure in Section~\ref{PREFfourexample}.

Now, we turn to a qualified version of preferential consequence.
It captures the idea that the contradictions in the conclusions should be rejected.

\begin{definition} \label{PREFdefprefdisrel}
Suppose $\cal L$ is a language, $\neg$ a unary connective of $\cal L$,
$\cal F$ the set of all wffs of $\cal L$, $\langle {\cal F}, {\cal V}, \models \rangle$
a semantic structure, and $\mid\!\sim$ a relation on ${\cal P}({\cal F}) \times {\cal F}$.
\\
We say that $\mid\!\sim$ is a {\it preferential-discriminative consequence relation}
iff there is a coherent choice function $\mu$ from $\bf D$ to ${\cal P}({\cal V})$
such that $\forall \: \Gamma \subseteq {\cal F}$, $\forall \: \alpha \in {\cal F}$,
$$\Gamma \mid\!\sim \alpha\; \textrm{iff}\; \mu(\M{\Gamma}) \subseteq \M{\alpha}\; \textrm{and}\;
\mu(\M{\Gamma}) \not\subseteq \M{\neg\alpha}.$$
In addition, if $\mu$ is LM, DP, etc., then so is $\mid\!\sim$.
\end{definition}
If a classical semantic structure is considered, the discriminative version does not bring something really new.
Indeed, the only difference will be to conclude nothing instead of everything in the face of inconsistent information.
On the other hand, with a many-valued structure,
the conclusions are rational even from inconsistent information.
The discriminative version will then reject the contradictions in the conclusions,
rendering the latter all the more rational.

In Definitions~\ref{PREFdefprefrel} and \ref{PREFdefprefdisrel}, the domain of the choice function is $\bf D$.
This is natural as only the elements of $\bf D$ play a role in the definition of a preferential(-discriminative) consequence relation. This point of view has been adopted in e.g. \cite{Lehmann2} (see Section~6).
Now, one might want a definition with choice functions of which the domain is ${\cal P}({\cal V})$.
In fact, some families of relations can be defined equivalently with $\bf D$ or ${\cal P}({\cal V})$.
For instance, as is noted in \cite{Lehmann2}, if $\mu$ is a coherent choice function from $\bf D$ to ${\cal P}({\cal V})$,
then the function $\mu'$ from ${\cal P}({\cal V})$ to ${\cal P}({\cal V})$ defined by $\mu'(V) = V \cap \mu(\M{\T{V}})$
is a coherent choice function which agrees with $\mu$ on $\bf D$.

Several characterizations for preferential consequence relations can be found in the literature
(e.g. \cite{KrausLehmannMagidor1, LehmannMagidor1, Lehmann1, Lehmann2, Schlechta2, Schlechta3, Schlechta1, Schlechta5}).
In particular, we will recall (in Section~\ref{PREFsysP}) a characterization that involves the well-known
system $\bf P$ of \cite{KrausLehmannMagidor1}.

As said previously, in the light of Proposition \ref{PREFkarlProp}, preferential(-discriminative) consequence relations
could have been introduced equivalently with preference structures. We opted for coherent choice functions
for two reasons.
First, they give a clearer meaning. Indeed, properties like Coherence have simple intuitive justifications,
whilst preference structures contain ``states'',
but it is not perfectly clear what a state is in daily life. By the way, in \cite{KrausLehmannMagidor1},
Kraus, Lehmann, and Magidor did not consider preference structures to be ontological
justifications for their interest in the formal systems investigated, but to be technical tools
to study those systems and in particular settle questions of interderivability and find efficient
decision procedures (see the end of Section~1.2 of \cite{KrausLehmannMagidor1}).

Second, in the proofs, we will work directly with choice functions and their properties,
not with preference structures. By the way, the techniques developed in the present paper
(especially in the discriminative case) can certainly be adapted to new properties.

\subsection{The system $\bf P$} \label{PREFsysP}

Gabbay, Makinson, Kraus, Lehmann, and Magidor
investigated extensively properties which should be satisfied 
by plausible non-monotonic consequence relations \cite{Gabbay1, Makinson5, Makinson4, KrausLehmannMagidor1, LehmannMagidor1}.
A certain set of properties, called the system $\bf P$, plays a central role in this area.
It is essentially due to Kraus, Lehmann, and Magidor \cite{KrausLehmannMagidor1} and has been investigated further in \cite{LehmannMagidor1}.
Let's present it.

\begin{definition}
Suppose $\cal L$ is a language containing the usual connectives $\neg$ and $\vee$,
$\cal F$ the set of all wffs of $\cal L$, $\langle {\cal F}, {\cal V}, \models \rangle$
a semantic structure, and $\mid\!\sim$ a relation on ${\cal F} \times {\cal F}$.
\\
Then, the system $\bf P$ is the set of the six following conditions:
$\forall \: \alpha, \beta, \gamma \in {\cal F}$,
\begin{description}
\item[Reflexivity] $\alpha \mid\!\sim \alpha$
\item[Left Logical Equivalence] $\begin{array}{c}
\underline{\vdash \alpha \leftrightarrow \beta \quad \alpha \mid\!\sim \gamma}\\
\beta \mid\!\sim \gamma \end{array}$
\item[Right Weakening] $\begin{array}{c}
\underline{\vdash \alpha \rightarrow \beta \quad \gamma \mid\!\sim \alpha}\\
\gamma \mid\!\sim \beta \end{array}$
\item[Cut] $\begin{array}{c}
\underline{\alpha \wedge \beta \mid\!\sim \gamma \quad \alpha \mid\!\sim \beta}\\
\alpha \mid\!\sim \gamma \end{array}$
\item[Cautious Monotonicity] $\begin{array}{c}
\underline{\alpha \mid\!\sim \beta \quad \alpha \mid\!\sim \gamma}\\
\alpha \wedge \beta \mid\!\sim \gamma \end{array}$
\item[Or] $\begin{array}{c}
\underline{\alpha \mid\!\sim \gamma \quad \beta \mid\!\sim \gamma}\\
\alpha \vee \beta \mid\!\sim \gamma \end{array}$
\end{description}
\end{definition}
Note that $\alpha \wedge \beta$ is a shorthand for $\neg(\neg\alpha \vee \neg\beta)$.
Similarly, $\alpha \rightarrow \beta$ and $\alpha \leftrightarrow \beta$ are shorthands.
Note again that $\bf P$ without $\bf Or$ is called $\bf C$. The system $\bf C$ is closely related to the
cumulative inference which was investigated by Makinson in \cite{Makinson5}.
In addition, it seems to correspond to what Gabbay proposed in \cite{Gabbay1}.
Concerning the rule $\bf Or$, it corresponds to the axiom CA of conditional logic.

All the properties in $\bf P$ are sound if we read $\alpha \mid\!\sim \beta$ as
``$\beta$ is a plausible consequence of $\alpha$''. In addition, $\bf P$ is complete
in the sense that it characterizes those consequence relations that
can be defined by a smooth transitive irreflexive preference structure.
This is what makes $\bf P$ central. More formally:

\begin{definition}
Suppose $\langle {\cal F}, {\cal V}, \models \rangle$ is a semantic structure.
\\
Then, ${\bf D}_f := \lbrace V \subseteq {\cal V} : \exists \: \alpha \in {\cal F}$, $V = \M{\alpha} \rbrace$.
\\
Suppose $\cal L$ is a language containing the usual connectives $\neg$ and $\vee$, and
$\cal F$ the set of all wffs of $\cal L$.
\\
Then define the following condition: $\forall \: v \in {\cal V}$, $\forall \: \alpha, \beta \in {\cal F}$,
$\forall \: \Gamma \subseteq {\cal F}$,
\begin{description}
\item[$(KLM0)$] $v \models \neg\alpha$ iff $v \not\models \alpha$;
\item[$(KLM1)$] $v \models \alpha \vee \beta$ iff $v \models \alpha$ or $v \models \beta$.
\item[$(KLM2)$] if for every finite subset $\Delta$ of $\Gamma$, $\M{\Delta} \not= \emptyset$,
then $\M{\Gamma} \not= \emptyset$.
\end{description}
\end{definition}
Note that $(KLM2)$ is called ``assumption of compactness'' in \cite{KrausLehmannMagidor1}.

\begin{proposition}\label{PREFKLM} {\bf \cite{KrausLehmannMagidor1}}~
Suppose $\cal L$ is a language containing the usual connectives $\neg$ and $\vee$, $\cal F$ the set of all wffs of $\cal L$,
$\langle {\cal F}, {\cal V}, \models \rangle$ a semantic structure satisfying $(KLM0)$--$(KLM2)$,
and $\mid\!\sim$ a relation of ${\cal F} \times {\cal F}$.
\\
Then, $\mid\!\sim$ satisfies all the properties of $\bf P$ iff there exists a ${\bf D}_f$-smooth transitive irreflexive
preference structure $\cal R$ on $\cal V$ such that $\forall \: \alpha, \beta \in {\cal F}$,
$\alpha \mid\!\sim \beta$ iff $\mu_{\cal R}(\M{\alpha}) \subseteq \M{\beta}$.
\end{proposition}
Note that $\mid\!\sim$ is a relation on ${\cal F} \times {\cal F}$, not ${\cal P}({\cal F}) \times {\cal F}$.
This difference is crucial. Indeed, if we adapt the conditions of $\bf P$ in the obvious way to
relations on ${\cal P}({\cal F}) \times {\cal F}$ and if we replace ${\bf D}_f$ by $\bf D$ in
Proposition~\ref{PREFKLM}, then the latter does no longer hold. This negative result was shown
by Schlechta in \cite{Schlechta2}.

Now, by Propositions~\ref{PREFkarlProp} and \ref{PREFKLM}, we immediately get the following representation
theorem:

\begin{proposition}
Suppose Definition~\ref{PREFdefprefrel} (of preferential consequence relations) is adapted
in the obvious way to relations on ${\cal F} \times {\cal F}$ (essentially, replace $\bf D$ by ${\bf D}_f$),
$\cal L$ is a language containing the usual connectives $\neg$ and $\vee$, $\cal F$ the set of all wffs of $\cal L$, $\mid\!\sim$ a relation on ${\cal F} \times {\cal F}$,
and $\langle {\cal F}, {\cal V}, \models \rangle$ a semantic structure such that $(KLM0)$--$(KLM2)$ hold
and $\forall \: V, W \in {\bf D}_f$, $V \cup W \in {\bf D}_f$ and $V \cap W \in {\bf D}_f$.
\\
Then, LM preferential consequence relations are precisely those relations that satisfy the system $\bf P$.
\end{proposition}

\subsubsection{Example with a classical semantic structure} \label{PREFclassicalexample}

Let $\cal L$ be a classical propositional language of which the atoms are
$r$, $q$, and $p$.
Intuitively, $r$ means Nixon is a republican, $q$ means Nixon is a quaker, and $p$ means Nixon is a pacifist.
Let $\cal F$ be the set of all wffs of $\cal L$,
$\cal V$ the set of all classical two-valued valuations of $\cal L$, and $\models$
the classical satisfaction relation for these objects.
Then, ${\cal V}$ is the set of the 8 following valuations: $v_0$, $v_1$, $v_2$, $v_3$, $v_4$, $v_5$, $v_6$, and $v_7$,
which are defined in the obvious way by the following table:
$$
\begin{array}{|c||c|c|c|}
\hline
               & r & q & p \\
\hline v_0 & 0 & 0 & 0 \\
\hline v_1 & 0 & 0 & 1 \\
\hline v_2 & 0 & 1 & 0 \\
\hline v_3 & 0 & 1 & 1 \\
\hline v_4 & 1 & 0 & 0 \\
\hline v_5 & 1 & 0 & 1 \\
\hline v_6 & 1 & 1 & 0 \\
\hline v_7 & 1 & 1 & 1 \\
\hline
\end{array}
$$
Now, consider the class of all republicans and the class of all quakers.
Consider that a republican is normal iff he is not a pacifist and that a quaker is normal iff he is a pacifist.
And, consider that a valuation $v$ is more normal than a valuation $w$ from the point of view
of a class $C$ iff
\begin{itemize}
\item Nixon is an individual of $C$ in both $v$ and $w$;
\item Nixon is normal in $v$;
\item Nixon is not normal in $w$.
\end{itemize}
In the following graph, there is an arrow from a valuation $v$
to a valuation $w$ iff $v$ is more normal than $w$ from the point of view of some class:
\begin{center}
\epsfig{file=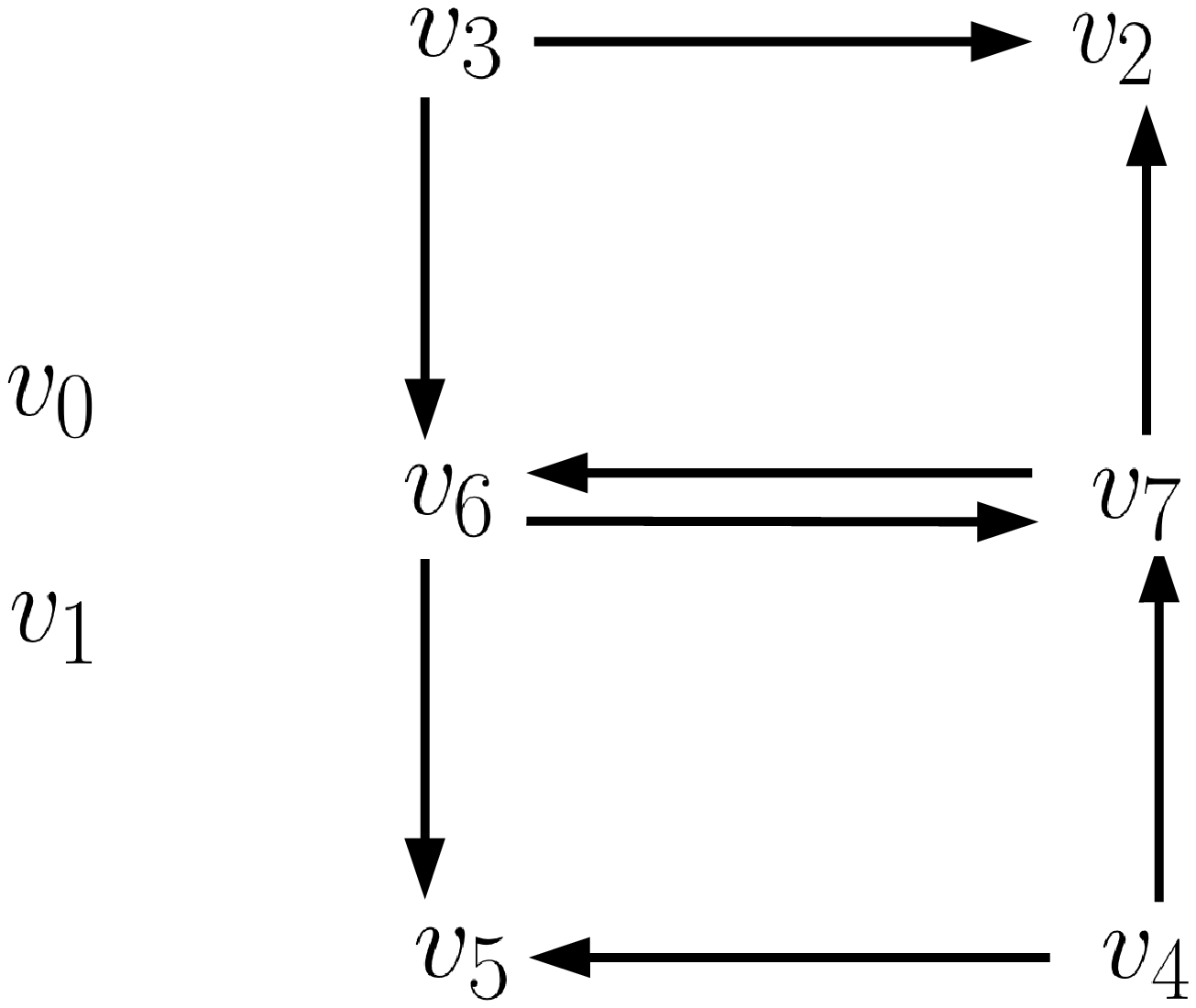, width=4cm}
\end{center}
Given those considerations a natural preference structure on $\cal V$ is
${\cal R} = \langle {\cal V}, l, \prec \rangle$, where
$l$ is identity and $\prec$ is the relation such that $\forall \: v, w \in {\cal V}$, we have
$v \prec w$ iff $(1)$ or $(2)$ below holds (i.e. there is an arrow from $v$ to $w$):
\begin{itemize}
\item[$(1)$] $v \models r$ and $v \models \neg p$ and $w \models r$ and $w \not\models \neg p$;
\item[$(2)$] $v \models q$ and $v \models p$ and $w \models q$ and $w \not\models p$.
\end{itemize}
Finally, let $\mid\!\sim$ be the preferential consequence relation defined by the
coherent choice function $\mu_{\cal R}$.

Then, $\mid\!\sim$ leads us to ``jump'' to plausible conclusions from incomplete information
and to revise previous ``hasty'' conclusions in the face of new and fuller information.
For instance, $r \mid\!\sim \neg p$ and $\lbrace r, p \rbrace \not\mid\!\sim \neg p$ and $q \mid\!\sim p$ and $\lbrace q, \neg p \rbrace \not\mid\!\sim p$.

However, $\mid\!\sim$ is not paraconsistent.
In addition, some sets of formulas are rendered useless,
because there is no preferred model for them, though there are models for them. For instance,
$\lbrace q, r \rbrace \mid\!\sim \alpha$, $\forall \: \alpha \in {\cal F}$.

\subsubsection{Example with the $\cal FOUR$ semantic structure} \label{PREFfourexample}

Consider the $\cal FOUR$ semantic structure $\langle {\cal F}_c, {\cal V}_4, \models_4 \rangle$
and suppose ${\cal A} = \lbrace r, q, p \rbrace$ (these objects have been defined
in Section~\ref{PREFfourframework}). In addition, make the same considerations
about Nixon, the classes, normality, etc., as in Section~\ref{PREFclassicalexample}, except that
this time a valuation $v$ is considered to be more normal than a valuation $w$
from the point of view of a class $C$ iff
\begin{itemize}
\item in both $v$ and $w$, the processor is informed that Nixon is an individual of $C$;
\item in $v$, he is informed that Nixon is normal and not informed of the contrary;
\item in $w$, he is not informed that Nixon is normal.
\end{itemize}
See Section~\ref{PREFfourframework} for recalls about the sources-processor systems.
Given those considerations a natural preference structure on ${\cal V}_4$ is
${\cal R} = \langle {\cal V}_4, l, \prec \rangle$, where
$l$ is identity and $\prec$ is the relation such that $\forall \: v, w \in {\cal V}_4$, we have
$v \prec w$ iff $(1)$ or $(2)$ below holds (i.e. $v$ is more normal than $w$ from the point of view of some class):
\begin{itemize}
\item[$(1)$] $v \models r$ and $v \models \neg p$ and $v \not\models p$ and $w \models r$ and $w \not\models \neg p$;
\item[$(2)$] $v \models q$ and $v \models p$ and $v \not\models \neg p$ and $w \models q$ and $w \not\models p$.
\end{itemize}
Let $\mid\!\sim$ be the preferential consequence relation defined by
the coherent choice function $\mu_{\cal R}$.

Then, again we ``jump'' to plausible conclusions
and revise previous ``hasty'' conclusions.
For instance, $r \mid\!\sim \neg p$ and $\lbrace r, p \rbrace \not\mid\!\sim \neg p$ and $q \mid\!\sim p$ and $\lbrace q, \neg p \rbrace \not\mid\!\sim p$.

In addition, $\mid\!\sim$ is paraconsistent. For instance,
$\lbrace p, \neg p, q \rbrace \mid\!\sim p$ and $\lbrace p, \neg p, q \rbrace \mid\!\sim \neg p$ and
$\lbrace p, \neg p, q \rbrace \mid\!\sim q$ and $\lbrace p, \neg p, q \rbrace \not\mid\!\sim \neg q$.
And, it happens less often that a set of formulas is rendered useless
because there is no preferred model for it, though there are models for it.
For instance, this time,
$\lbrace q, r \rbrace \mid\!\sim p$ and $\lbrace q, r \rbrace \mid\!\sim \neg p$ and $\lbrace q, r \rbrace \mid\!\sim q$ and $\lbrace q, r \rbrace \not\mid\!\sim \neg q$ and $\lbrace q, r \rbrace \mid\!\sim r$ and
$\lbrace q, r \rbrace \not\mid\!\sim \neg r$.

However, $\mid\!\sim$ does not satisfy the Disjunctive Syllogism.
Indeed, for instance, $\lbrace \neg r, r \vee q \rbrace \not\mid\!\sim q$.

\section{Contributions} \label{PREFcontribu}

The main contributions of the present paper are summarized below.
We characterized (in many cases, by purely syntactic conditions) families of preferential
and preferential-discriminative consequence relations.
Sometimes, we will need to make some assumptions about the semantic structure under consideration.
However, no assumption will be needed for the three following families:
\begin{itemize}
\item the preferential consequence relations (Section~\ref{PREFprefCR});
\item the DP preferential consequence relations (Section~\ref{PREFdpprefCR});
\item the DP LM preferential consequence relations (Section~\ref{PREFdpprefCR}).
\end{itemize}
We will assume $(A1)$ and $(A3)$ for:
\begin{itemize}
\item the CP preferential-discriminative consequence relations (Section~\ref{PREFprefdisCR});
\item the CP DP preferential-discriminative consequence relations (Section~\ref{PREFdpprefdisCR});
\item the CP DP LM preferential-discriminative consequence relations (Section~\ref{PREFdpprefdisCR}).
\end{itemize}
And, we will need $(A1)$, $(A2)$, and $(A3)$ for:
\begin{itemize}
\item the preferential-discriminative consequence relations (Section~\ref{PREFprefdisCR});
\item the DP preferential-discriminative consequence relations  (Section~\ref{PREFdpprefdisCR});
\item the DP LM preferential-discriminative consequence relations (Section~\ref{PREFdpprefdisCR}).
\end{itemize}

\subsection{The non-discriminative and definability preserving case} \label{PREFdpprefCR}

The characterizations in this section have already been given in Proposition 3.1 of \cite{Schlechta1},
under the assumption that a classical propositional semantic structure is considered.
Using the same techniques as those of Schlechta,
we show easily that his characterizations hold with any semantic structure.

\begin{notation}
Let $\langle {\cal F}, {\cal V}, \models \rangle$ be a semantic structure and
$\mid\!\sim$ a relation on ${\cal P}({\cal F}) \times {\cal F}$.
\\
Then, consider the following conditions: $\forall \: \Gamma, \Delta \subseteq {\cal F}$,
\begin{description}
\item[$(\mid\!\sim$$0)$] if $\C{\vdash}{\Gamma} = \C{\vdash}{\Delta}$, then $\C{\mid\!\sim}{\Gamma} = \C{\mid\!\sim}{\Delta}$;
\item[$(\mid\!\sim$$1)$] $\C{\vdash}{\C{\mid\!\sim}{\Gamma}} = \C{\mid\!\sim}{\Gamma}$;
\item[$(\mid\!\sim$$2)$] $\Gamma \subseteq \C{\mid\!\sim}{\Gamma}$;
\item[$(\mid\!\sim$$3)$] $\CC{\mid\!\sim}{\Gamma}{\Delta} \subseteq
\CC{\vdash}{\C{\mid\!\sim}{\Gamma}}{\Delta}$;
\item[$(\mid\!\sim$$4)$] if $\Gamma \subseteq \C{\vdash}{\Delta} \subseteq \C{\mid\!\sim}{\Gamma}$,
then $\C{\mid\!\sim}{\Gamma} \subseteq \C{\mid\!\sim}{\Delta}$.
\end{description}
\end{notation}
Note that those conditions are purely syntactic
when there is a proof system available for $\vdash$
(which is the case with e.g. the classical, $\cal FOUR$, and $J_3$ semantic structures).

\begin{proposition} \label{PREFrepClaSyn}
Let ${\cal S} = \langle {\cal F}, {\cal V}, \models \rangle$ be a semantic structure and
$\mid\!\sim$ a relation on ${\cal P}({\cal F}) \times {\cal F}$. Then,
\begin{description}
\item[$(0)$] $\mid\!\sim$ is a DP preferential consequence relation iff $(\mid\!\sim$$0)$, $(\mid\!\sim$$1)$, $(\mid\!\sim$$2)$, and
$(\mid\!\sim$$3)$ hold;
\item[$(1)$] $\mid\!\sim$ is a DP LM preferential consequence relation iff $(\mid\!\sim$$0)$, $(\mid\!\sim$$1)$, $(\mid\!\sim$$2)$,
$(\mid\!\sim$$3)$, and $(\mid\!\sim$$4)$ hold.
\end{description}
\end{proposition}

\begin{proof}
{\it Proof of $(0)$.} Direction: ``$\rightarrow$''.
\\
By hypothesis, there exists a DP coherent choice function $\mu$ from $\bf D$ to ${\cal P}({\cal V})$
such that $\forall \: \Gamma \subseteq {\cal F}$,
\\
$\C{\mid\!\sim}{\Gamma} = \T{\mu(\M{\Gamma})}$.
We will show:
\\
$(0.0)$\quad $\mid\!\sim$ satisfies $(\mid\!\sim$$0)$;
\\
$(0.1)$\quad $\mid\!\sim$ satisfies $(\mid\!\sim$$1)$;
\\
$(0.2)$\quad $\mid\!\sim$ satisfies $(\mid\!\sim$$2)$.
\\
Before turning to $(\mid\!\sim$$3)$, we need a preliminary result:
\\
$(0.3)$\quad $\forall \: \Gamma \subseteq {\cal F}$, we have $\mu(\M{\Gamma}) = \M{\C{\mid\!\sim}{\Gamma}}$;
\\
$(0.4)$\quad $\mid\!\sim$ satisfies $(\mid\!\sim$$3)$.

Direction: ``$\leftarrow$''.
\\
Suppose $\mid\!\sim$ satisfies
$(\mid\!\sim$$0)$, $(\mid\!\sim$$1)$, $(\mid\!\sim$$2)$, and $(\mid\!\sim$$3)$.
\\
Let $\mu$ be the function from $\bf D$ to ${\cal P}({\cal V})$ such that
$\forall \: \Gamma \subseteq {\cal F}$, $\mu(\M{\Gamma}) = \M{\C{\mid\!\sim}{\Gamma}}$.
\\
Then, $\mu$ is well-defined.
\\
Indeed, If $\Gamma, \Delta \subseteq {\cal F}$ and
$\M{\Gamma} = \M{\Delta}$, then
$\C{\vdash}{\Gamma} = \C{\vdash}{\Delta}$, thus, by $(\mid\!\sim$$0)$,
$\C{\mid\!\sim}{\Gamma} = \C{\mid\!\sim}{\Delta}$.
\\
In addition, $\mu$ is obviously DP.
We show the following which ends the proof:
\\
$(0.5)$ $\mu$ is a choice function;
\\
$(0.6)$ $\mu$ is coherent;
\\
$(0.7)$ $\forall \: \Gamma \subseteq {\cal F}$, we have $\C{\mid\!\sim}{\Gamma} = \T{\mu(\M{\Gamma})}$.
\\ \\
{\it Proof of $(0.0)$.} Let $\Gamma, \Delta \subseteq {\cal F}$ and suppose
$\C{\vdash}{\Gamma} = \C{\vdash}{\Delta}$.
\\
Then, $\M{\Gamma} = \M{\Delta}$. Thus,
$\C{\mid\!\sim}{\Gamma} = \T{\mu(\M{\Gamma})} = \T{\mu(\M{\Delta})} = \C{\mid\!\sim}{\Delta}$.
\\ \\
{\it Proof of $(0.1)$.} Let $\Gamma \subseteq {\cal F}$.
Then, $\C{\vdash}{\C{\mid\!\sim}{\Gamma}} = \C{\vdash}{\T{\mu(\M{\Gamma})}}
= \T{\M{\T{\mu(\M{\Gamma})}}}
= \C{\mid\!\sim}{\Gamma}$.
\\ \\
{\it Proof of $(0.2)$.} Let $\Gamma \subseteq {\cal F}$.
Then, $\Gamma \subseteq \T{\M{\Gamma}} \subseteq \T{\mu(\M{\Gamma})} =  \C{\mid\!\sim}{\Gamma}$.
\\ \\
{\it Proof of $(0.3)$.} Let $\Gamma \subseteq {\cal F}$. As, $\mu$ is DP, $\mu(\M{\Gamma}) \in {\bf D}$.
\\
Thus, $\exists \: \Gamma' \subseteq {\cal F}$, $\mu(\M{\Gamma}) = \M{\Gamma'}$.
Therefore, $\mu(\M{\Gamma}) = \M{\Gamma'} =
\M{\T{\M{\Gamma'}}} = \M{\T{\mu(\M{\Gamma})}} = \M{\C{\mid\!\sim}{\Gamma}}$.
\\ \\
{\it Proof of $(0.4)$.} Let $\Gamma, \Delta \subseteq {\cal F}$.
As, $\MM{\Gamma}{\Delta} \subseteq \M{\Gamma}$ and $\mu$ is coherent,
$\mu(\M{\Gamma}) \cap \MM{\Gamma}{\Delta} \subseteq \mu(\MM{\Gamma}{\Delta})$.
\\
Therefore,
$\CC{\mid\!\sim}{\Gamma}{\Delta} = \T{\mu(\MM{\Gamma}{\Delta})} \subseteq
\T{\mu(\M{\Gamma}) \cap \MM{\Gamma}{\Delta}} = \T{\mu(\M{\Gamma}) \cap \M{\Delta}}$.
\\
Thus, by $(0.0)$,
$\CC{\mid\!\sim}{\Gamma}{\Delta} \subseteq \T{\M{\C{\mid\!\sim}{\Gamma}} \cap \M{\Delta}} =
 \T{\MM{\C{\mid\!\sim}{\Gamma}}{\Delta}} =
\CC{\vdash}{\C{\mid\!\sim}{\Gamma}}{\Delta}$.
\\ \\
{\it Proof of $(0.5)$.} Let $\Gamma \subseteq {\cal F}$. Then,
$\mu(\M{\Gamma}) = \M{\C{\mid\!\sim}{\Gamma}}$, which is, by $(\mid\!\sim$$2)$,
a subset of $\M{\Gamma}$.
\\ \\
{\it Proof of $(0.6)$.} Let $\Gamma, \Delta \subseteq {\cal F}$ and suppose
$\M{\Gamma} \subseteq \M{\Delta}$.
\\
Then, $\mu(\M{\Delta}) \cap \M{\Gamma} =
\M{\C{\mid\!\sim}{\Delta}} \cap \M{\Gamma} = \MM{\C{\mid\!\sim}{\Delta}}{\Gamma}$.
\\
But, by $(\mid\!\sim$$3)$, $\MM{\C{\mid\!\sim}{\Delta}}{\Gamma} \subseteq
\M{\CC{\mid\!\sim}{\Delta}{\Gamma}} = \mu(\MM{\Delta}{\Gamma}) = \mu(\M{\Gamma})$.
\\ \\
{\it Proof of $(0.7)$.} Let $\Gamma \subseteq {\cal F}$. Then, by $(\mid\!\sim$$1)$,
$\C{\mid\!\sim}{\Gamma} = \C{\vdash}{\C{\mid\!\sim}{\Gamma}} =
\T{\M{\C{\mid\!\sim}{\Gamma}}} = \T{\mu(\M{\Gamma})}$.
\\ \\
{\it Proof of $(1)$.} Direction: ``$\rightarrow$''.
\\
Verbatim the same proof as for $(0)$, except that in addition $\mu$ is LM.
\\
We use it to show that $\mid\!\sim$ satisfies $(\mid\!\sim$$4)$.
\\
Let $\Gamma, \Delta \subseteq {\cal F}$ and suppose
$\Gamma \subseteq \C{\vdash}{\Delta} \subseteq \C{\mid\!\sim}{\Gamma}$.
\\
Then, by $(0.3)$, $\mu(\M{\Gamma}) = \M{\C{\mid\!\sim}{\Gamma}}
\subseteq \M{\C{\vdash}{\Delta}} = \M{\Delta} \subseteq \M{\Gamma}$.
\\
Therefore, as $\mu$ is locally monotonic, $\mu(\M{\Delta}) \subseteq \mu(\M{\Gamma})$.
\\
Thus, $\C{\mid\!\sim}{\Gamma} = \T{\mu(\M{\Gamma})} \subseteq \T{\mu(\M{\Delta})} = \C{\mid\!\sim}{\Delta}$.

Direction: ``$\leftarrow$''.
\\
Verbatim the same proof as for $(0)$, except that in addition $(\mid\!\sim$$4)$ is satisfied.
\\
We use it to show that $\mu$ is locally monotonic.
\\
Let $\Gamma, \Delta \subseteq {\cal F}$ and suppose
$\mu(\M{\Gamma}) \subseteq \M{\Delta} \subseteq \M{\Gamma}$.
\\
Then, $\M{\C{\mid\!\sim}{\Gamma}} \subseteq \M{\Delta} \subseteq \M{\Gamma}$.
Therefore, $\Gamma \subseteq \T{\M{\Gamma}} \subseteq \T{\M{\Delta}} = \C{\vdash}{\Delta}$.
\\
On the other hand, $\C{\vdash}{\Delta} = \T{\M{\Delta}}
\subseteq \T{\M{\C{\mid\!\sim}{\Gamma}}} = \C{\vdash}{\C{\mid\!\sim}{\Gamma}}$
which is, by $(\mid\!\sim$$1)$, equal to $\C{\mid\!\sim}{\Gamma}$.
\\
Thus, by $(\mid\!\sim$$4)$, we have $\C{\mid\!\sim}{\Gamma} \subseteq \C{\mid\!\sim}{\Delta}$.
Therefore, $\mu(\M{\Delta}) = \M{\C{\mid\!\sim}{\Delta}} \subseteq \M{\C{\mid\!\sim}{\Gamma}} = \mu(\M{\Gamma})$.\qed
\end{proof}

\subsection{The non-discriminative and not necessarily definability preserving case} \label{PREFprefCR}

In this section, we will characterize the family of all preferential consequence relations.
Unlike in Section~\ref{PREFdpprefCR}, our conditions will not be purely syntactic (i.e. using only $\vdash$, $\mid\!\sim$, etc.). In fact, properties like Coherence cannot be translated in syntactic terms
because the choice functions under consideration are not necessarily definability preserving.
Indeed, we do no longer have at our disposal the
equality: $\mu(\M{\Gamma}) = \M{\C{\mid\!\sim}{\Gamma}}$, which is of great help to perform the translation and which holds precisely because of Definability Preservation.

In Proposition 5.2.11 of \cite{Schlechta5}, K. Schlechta provided a characterization of
the aforementioned family, under the assumption
that a classical propositional semantic structure is considered.
Note that most of his work is done in a very general, in fact algebraic, framework.
Only at the end, he applied his general lemmas in a classical framework to
get the characterization.
The conditions he gave, as ours, are not purely syntactic (e.g. they involve the notion of model, etc.).
Moreover, some limits of what can be done in this area have been put in evidence by Schlechta.
Approximatively, he showed in Proposition 5.2.15 of the same book that,
in an infinite classical framework, there does not exist a characterization containing only
conditions which are universally quantified, of limited size, and using
only simple operations (like e.g. $\cup$, $\cap$, $\setminus$).

The purpose of the present section is to provided a new characterization,
more elegant than the one of Schlechta and that hold with any semantic structure.
To do so, we have been inspired by the algebraic part of the work of Schlechta
(see Proposition~5.2.5 of \cite{Schlechta5}).
Technically, the idea begins by building from any function $f$, a coherent choice function $\mup{f}$ such that
whenever $f$ ``covers'' some coherent choice function, it necessarily covers $\mup{f}$.

\begin{definition}
Let $\cal V$ be a set, ${\bf V}$ and ${\bf W}$ subsets of ${\cal P}({\cal V})$,
and $f$ a function from ${\bf V}$ to ${\bf W}$.
\\
We denote by $\mup{f}$ the function from ${\bf V}$ to ${\cal P}({\cal V})$ such that $\forall \: V \in {\bf V}$,
$$
\mupp{f}{V} = \lbrace v \in V : \forall \: W \in {\bf V},\; \textrm{if}\; v \in W \subseteq V,\; \textrm{then}\; v \in f(W) \rbrace.
$$
\end{definition}

\begin{lemma} \label{PREFmufestcf}
Let $\cal V$ be a set, ${\bf V}$ and ${\bf W}$ subsets of ${\cal P}({\cal V})$,
and $f$ a function from ${\bf V}$ to ${\bf W}$.
\\
Then, $\mup{f}$ is a coherent choice function.
\end{lemma}

\begin{proof} $\mup{f}$ is obviously a choice function.
It remains to show that it is coherent.
\\
Suppose $V, W \in {\bf V}$,
$V \subseteq W$, and $v \in \mupp{f}{W} \cap V$. We show $v \in \mupp{f}{V}$.
\\
To do so, suppose the contrary, i.e. suppose $v \not\in \mupp{f}{V}$.
\\
Then, as $v \in V$, we have
$\exists \: Z \in {\bf V}$, $Z \subseteq V$, $v \in Z$, and $v \not\in f(Z)$.
\\
But, $V \subseteq W$, thus $Z \subseteq W$. Therefore, by definition of $\mup{f}$,
$v \not\in \mupp{f}{W}$, which is impossible.\qed
\end{proof}

\begin{lemma}\label{PREFprefGen}
Let $\cal V$ be a set, ${\bf V}$, ${\bf W}$, and $\bf X$ subsets of ${\cal P}({\cal V})$,
$f$ a function from ${\bf V}$ to ${\bf W}$, and $\mu$ a coherent choice function from $\bf V$ to $\bf X$
such that $\forall \: V \in {\bf V}$, $f(V) = \M{\T{\mu(V)}}$.
\\
Then, $\forall \: V \in {\bf V}$, $f(V) = \M{\T{\mupp{f}{V}}}$.
\end{lemma}

\begin{proof} Let $V \in {\bf V}$. We show $f(V) = \M{\T{\mupp{f}{V}}}$.
\\
Case~1: $\exists \: v \in \mu(V)$, $v \not\in \mupp{f}{V}$.
\\
As $\mu(V) \subseteq V$, we have $v \in V$.
\\
Thus, by definition of $\mup{f}$,
$\exists \: W \in {\bf V}$, $W \subseteq V$, $v \in W$, and $v \not\in f(W) = \M{\T{\mu(W)}}
\supseteq \mu(W)$.
\\
On the other hand, as $\mu$ is coherent, 
$\mu(V) \cap W \subseteq \mu(W)$. Thus, $v \in \mu(W)$, which is impossible.
\\
Case~2: $\mu(V) \subseteq \mupp{f}{V}$.
\\
Case~2.1: $\exists \: v \in \mupp{f}{V}$, $v \not\in f(V)$.
\\
Then, $\exists \: W \in {\bf V}$, $W \subseteq V$, $v \in W$, and $v \not\in f(W)$.
Indeed, just take $V$ itself for the choice of $W$.
\\
Therefore,
$v \not \in \mupp{f}{V}$, which is impossible.
\\
Case~2.2: $\mupp{f}{V} \subseteq f(V)$.
\\
Then, $f(V) = \M{\T{\mu(V)}} \subseteq \M{\T{\mupp{f}{V}}} \subseteq \M{\T{f(V)}} =
\M{\T{\M{\T{\mu(V)}}}} = \M{\T{\mu(V)}} = f(V)$.\qed
\end{proof}
Now, everything is ready to show the representation result.

\begin{notation}
Let $\langle {\cal F}, {\cal V}, \models \rangle$ be a semantic structure and
$\mid\!\sim$ a relation on ${\cal P}({\cal F}) \times {\cal F}$.
\\
Then, consider the following condition: $\forall \: \Gamma \subseteq {\cal F}$,
\begin{description}
\item[$(\mid\!\sim$$5)$] $\C{\mid\!\sim}{\Gamma} = \T{\lbrace v \in \M{\Gamma} : \forall \: \Delta \subseteq {\cal F}$,
if $v \in \M{\Delta} \subseteq \M{\Gamma}$, then $v \in \M{\C{\mid\!\sim}{\Delta}} \rbrace}$.
\end{description}
\end{notation}

\begin{proposition} \label{PREFrepGen}
Let $\langle {\cal F}, {\cal V}, \models \rangle$ be a semantic structure and
$\mid\!\sim$ a relation on ${\cal P}({\cal F}) \times {\cal F}$.
\\
Then, $\mid\!\sim$ is a preferential consequence relation iff $(\mid\!\sim$$5)$ holds.
\end{proposition}

\begin{proof}
{\it Direction:~``$\rightarrow$''.}
\\
There exists a coherent choice function $\mu$ from
$\bf D$ to ${\cal P}({\cal V})$ such that $\forall \: \Gamma \subseteq {\cal F}$,
$\C{\mid\!\sim}{\Gamma} = \T{\mu(\M{\Gamma})}$.
\\
Let $f$ be the function from $\bf D$ to $\bf D$ such that
$\forall \: V \in {\bf D}$, we have $f(V) = \M{\T{\mu(V)}}$.
\\
By Lemma~\ref{PREFprefGen}, $\forall \: V \in {\bf D}$, we have $f(V) = \M{\T{\mupp{f}{V}}}$.
\\
Note that $\forall \: \Gamma \subseteq {\cal F}$, $f(\M{\Gamma}) = \M{\T{\mu(\M{\Gamma})}} =
\M{\C{\mid\!\sim}{\Gamma}}$.
\\
We show that $(\mid\!\sim$$5)$ holds.
Let $\Gamma \subseteq {\cal F}$.
\\
Then,
$\C{\mid\!\sim}{\Gamma} =
\T{\mu(\M{\Gamma})} =
\T{\M{\T{\mu(\M{\Gamma})}}} =
\T{f(\M{\Gamma})} =
\T{\M{\T{\mupp{f}{\M{\Gamma}}}}} =
\T{\mupp{f}{\M{\Gamma}}} =
\\
\T{\lbrace v \in \M{\Gamma} : \forall \: W \in {\bf D}$,
if $v \in W \subseteq \M{\Gamma}$, then $v \in f(W) \rbrace} =
\\
\T{\lbrace v \in \M{\Gamma} : \forall \: \Delta \subseteq {\cal F}$,
if $v \in \M{\Delta} \subseteq \M{\Gamma}$, then $v \in f(\M{\Delta}) \rbrace} =
\\
\T{\lbrace v \in \M{\Gamma} : \forall \: \Delta \subseteq {\cal F}$,
if $v \in \M{\Delta} \subseteq \M{\Gamma}$, then $v \in \M{\C{\mid\!\sim}{\Delta}} \rbrace}$.
\\ \\
{\it Direction:~``$\leftarrow$''.}
\\
Suppose $\mid\!\sim$ satisfies $(\mid\!\sim$$5)$.
\\
Let $f$ be the function from $\bf D$ to $\bf D$ such that
$\forall \: \Gamma \subseteq {\cal F}$, we have $f(\M{\Gamma}) = \M{\C{\mid\!\sim}{\Gamma}}$.
\\
Note that $f$ is well-defined. Indeed, 
if $\Gamma, \Delta \subseteq {\cal F}$ and $\M{\Gamma} = \M{\Delta}$, then, by $(\mid\!\sim$$5)$,
$\C{\mid\!\sim}{\Gamma} = \C{\mid\!\sim}{\Delta}$.
\\
In addition, by $(\mid\!\sim$$5)$, we clearly have
$\forall \: \Gamma \subseteq {\cal F}$, $\C{\mid\!\sim}{\Gamma} = \T{\mupp{f}{\M{\Gamma}}}$.
\\
And finally, by Lemma~\ref{PREFmufestcf}, $\mup{f}$ is a coherent choice function.\qed
\end{proof}

\subsection{The discriminative and definability preserving case} \label{PREFdpprefdisCR}

In this section, we will characterize certain families of DP preferential-discriminative consequence relations.
To do so, we will develop new techniques (especially Lemmas~\ref{PREF3sim} and \ref{PREFPrf2ConArgSyn} below).
We need basic notations and an inductive construction:

\begin{notation}\label{PREFNZect}
$\mathbb{N}$ denotes the natural numbers including 0: $\lbrace 0, 1, 2, \ldots, \rbrace$.
\\
$\mathbb{N}^{+}$ denotes the strictly positive natural numbers: $\lbrace 1, 2, \ldots, \rbrace$.
\\
$\mathbb{Z}$ denotes the integers.
\\
Let $i, j \in \mathbb{Z}$. Then, $[i, j]$ denotes the set of all $k \in \mathbb{Z}$ such that $i \leq k \leq j$.
\\
Let $\cal L$ be a language, $\vee$ a binary connective of $\cal L$, $\cal F$ the set of all wffs of $\cal L$,
and $\beta_1, \beta_2, \ldots, \beta_r \in {\cal F}$.
\\
Whenever we write $\beta_1 \vee \beta_2 \vee \ldots \vee \beta_r$,
we mean $( \ldots (( \beta_1 \vee \beta_2 ) \vee \beta_3 ) \vee \ldots \vee \beta_{r-1} ) \vee \beta_r$.
\end{notation}

\begin{definition}
Let $\cal L$ be a language, $\neg$ a unary connective of $\cal L$, $\cal F$ the set of all wffs of $\cal L$,
$\langle {\cal F}, {\cal V}, \models \rangle$ a semantic structure, $\mid\!\sim$ a relation on
${\cal P}({\cal F}) \times {\cal F}$, and $\Gamma \subseteq {\cal F}$. Then,
$$\monHi{1}{\Gamma} := \lbrace \neg\beta \in {\cal F} :
\beta \in \CC{\vdash}{\Gamma}{\C{\mid\!\sim}{\Gamma}} \setminus \C{\mid\!\sim}{\Gamma} \; \textrm{and}\;
\neg\beta \not\in \CC{\vdash}{\Gamma}{\C{\mid\!\sim}{\Gamma}}\rbrace.$$
Let $i \in \mathbb{N}$ with $i \geq 2$. Then,
$$\monHi{i}{\Gamma} := \lbrace \neg\beta \in {\cal F} :
\left \{ \begin{array}{l}
\beta \in \CCC{\vdash}{\Gamma}{\C{\mid\!\sim}{\Gamma}}
{\monHi{1}{\Gamma}, \ldots, \monHi{i-1}{\Gamma}} \setminus \C{\mid\!\sim}{\Gamma}\; \textrm{and}\\
\neg\beta \not\in \CCC{\vdash}{\Gamma}{\C{\mid\!\sim}{\Gamma}}
{\monHi{1}{\Gamma}, \ldots, \monHi{i-1}{\Gamma}} \end{array} \right . \rbrace.$$
$$\monH{\Gamma} := \bigcup_{i \in \mathbb{N}^+} \monHi{i}{\Gamma}.$$
\end{definition}

\begin{definition}
Suppose $\cal L$ is a language, $\neg$ a unary connective of $\cal L$,
$\vee$ a binary connective of $\cal L$,
$\cal F$ the set of all wffs of $\cal L$,
$\langle {\cal F}, {\cal V}, \models \rangle$ a semantic structure, and $\mid\!\sim$ a relation on
${\cal P}({\cal F}) \times {\cal F}$.
\\
Then, consider the following conditions: $\forall \: \Gamma, \Delta \subseteq {\cal F}$, $\forall \: \alpha, \beta \in {\cal F}$,
\begin{description}
\item[$(\mid\!\sim$$6)$]
if $\beta \in \CC{\vdash}{\Gamma}{\C{\mid\!\sim}{\Gamma}} \setminus \C{\mid\!\sim}{\Gamma}$ and
$\neg\alpha \in \CCC{\vdash}{\Gamma}{\C{\mid\!\sim}{\Gamma}}{\neg\beta}$, then
$\alpha \not\in \C{\mid\!\sim}{\Gamma}$;
\item[$(\mid\!\sim$$7)$]
if $\alpha \in \CC{\vdash}{\Gamma}{\C{\mid\!\sim}{\Gamma}} \setminus \C{\mid\!\sim}{\Gamma}$ and
$\beta \in \CCC{\vdash}{\Gamma}{\C{\mid\!\sim}{\Gamma}}{\neg\alpha} \setminus \C{\mid\!\sim}{\Gamma}$, then
$\alpha \vee \beta \not\in \C{\mid\!\sim}{\Gamma}$;
\item[$(\mid\!\sim$$8)$]
if $\alpha \in \C{\mid\!\sim}{\Gamma}$, then $\neg\alpha \not\in \CC{\vdash}{\Gamma}{\C{\mid\!\sim}{\Gamma}}$;
\item[$(\mid\!\sim$$9)$]
if $\Delta \subseteq \C{\vdash}{\Gamma}$, then
$\C{\mid\!\sim}{\Gamma} \cup \monH{\Gamma} \subseteq \CCCC{\vdash}{\Delta}{\C{\mid\!\sim}{\Delta}}{\monH{\Delta}}{\Gamma}$;
\item[$(\mid\!\sim$$10)$]
if $\Gamma \subseteq \C{\vdash}{\Delta} \subseteq
\CCC{\vdash}{\Gamma}{\C{\mid\!\sim}{\Gamma}}{\monH{\Gamma}}$, then
$\C{\mid\!\sim}{\Gamma} \cup \monH{\Gamma} \subseteq \CCC{\vdash}{\Delta}{\C{\mid\!\sim}{\Delta}}{\monH{\Delta}}$;
\item[$(\mid\!\sim$$11)$]
if $\Gamma$ is consistent, then $\C{\mid\!\sim}{\Gamma}$ is consistent,
$\Gamma \subseteq \C{\mid\!\sim}{\Gamma}$, and
$\C{\vdash}{\C{\mid\!\sim}{\Gamma}} = \C{\mid\!\sim}{\Gamma}$.
\end{description}
\end{definition}
Note that those conditions are purely syntactic
when there is a proof system available for $\vdash$.

\begin{proposition} \label{PREFrepArgSyn}
Suppose $\cal L$ is a language, $\neg$ a unary connective of $\cal L$,
$\vee$ and $\wedge$ binary connectives of $\cal L$,
$\cal F$ the set of all wffs of $\cal L$,
$\langle {\cal F}, {\cal V}, \models \rangle$ a semantic structure satisfying $(A1)$ and $(A3)$, and $\mid\!\sim$ a relation on
${\cal P}({\cal F}) \times {\cal F}$. Then,
\begin{description}
\item[$(0)$] $\mid\!\sim$ is a CP DP preferential-discriminative consequence relation iff
$(\mid\!\sim$$0)$, $(\mid\!\sim$$6)$, $(\mid\!\sim$$7)$, $(\mid\!\sim$$8)$, $(\mid\!\sim$$9)$, and $(\mid\!\sim$$11)$ hold;
\item[$(1)$] $\mid\!\sim$ is a CP DP LM preferential-discriminative consequence relation iff
$(\mid\!\sim$$0)$, $(\mid\!\sim$$6)$, $(\mid\!\sim$$7)$, $(\mid\!\sim$$8)$, $(\mid\!\sim$$9)$, $(\mid\!\sim$$10)$, and $(\mid\!\sim$$11)$ hold.
\end{description}
Suppose $\langle {\cal F}, {\cal V}, \models \rangle$ satisfies $(A2)$ too. Then,
\begin{description}
\item[$(2)$] $\mid\!\sim$ is a DP preferential-discriminative consequence relation iff
$(\mid\!\sim$$0)$, $(\mid\!\sim$$6)$, $(\mid\!\sim$$7)$, $(\mid\!\sim$$8)$, and $(\mid\!\sim$$9)$ hold;
\item[$(3)$] $\mid\!\sim$ is a DP LM preferential-discriminative consequence relation iff
$(\mid\!\sim$$0)$, $(\mid\!\sim$$6)$, $(\mid\!\sim$$7)$, $(\mid\!\sim$$8)$, $(\mid\!\sim$$9)$, and $(\mid\!\sim$$10)$ hold.
\end{description}
\end{proposition}
The proof of Proposition~\ref{PREFrepArgSyn} has been relegated at the end of Section~\ref{PREFdpprefdisCR}.
We need first Notation~\ref{PREFNZect},
Definition~\ref{PREFF} and Lemmas~\ref{PREFsensfacile}, \ref{PREF3sim}, and \ref{PREFPrf2ConArgSyn} below.
Here are some purely technical tools:

\begin{definition} \label{PREFF}
Suppose $\cal L$ is a language, $\neg$ a unary connective of $\cal L$,
$\vee$ a binary connective of $\cal L$,
$\cal F$ the set of all wffs of $\cal L$,
$\langle {\cal F}, {\cal V}, \models \rangle$ a semantic structure satisfying $(A1)$, $\mid\!\sim$ a relation on
${\cal P}({\cal F}) \times {\cal F}$, and $\Gamma \subseteq {\cal F}$. Then,
$$
\Mi{1}{\Gamma} := \lbrace v \in \MM{\Gamma}{\C{\mid\!\sim}{\Gamma}} : \exists \:
\beta \in \T{\MM{\Gamma}{\C{\mid\!\sim}{\Gamma}}} \setminus \C{\mid\!\sim}{\Gamma},\;
v \not\in \M{\neg\beta} \rbrace.
$$
Let $i \in \mathbb{N}$ with $i \geq 2$. Then,
$$\Mi{i}{\Gamma} := \lbrace v \in \MM{\Gamma}{\C{\mid\!\sim}{\Gamma}} \setminus \Mi{1}{\Gamma}
\cup \ldots \cup \Mi{i-1}{\Gamma} : \exists \: \beta \in
\T{\MM{\Gamma}{\C{\mid\!\sim}{\Gamma}} \setminus
\Mi{1}{\Gamma} \cup \ldots \cup \Mi{i-1}{\Gamma}} \setminus \C{\mid\!\sim}{\Gamma},\;
v \not\in \M{\neg\beta} \rbrace.
$$
$$
\Mp{\Gamma} := \bigcup_{i \in \mathbb{N}^+} \Mi{i}{\Gamma}
$$
$$
\n{\Gamma} := |\lbrace i \in \mathbb{N}^+ : \Mi{i}{\Gamma} \not= \emptyset \rbrace|
$$
Suppose $\Mi{1}{\Gamma} \not= \emptyset$.
Then, we denote by $\beti{1}{\Gamma}$ an element of $\cal F$, chosen arbitrarily,
such that
\\
$\exists \: r \in \mathbb{N}^+$,
$\exists \: v_1, v_2, \ldots, v_r \in {\cal V}$, and
$\exists \: \beta_1, \beta_2, \ldots, \beta_r \in {\cal F}$ with
$\Mi{1}{\Gamma} = \lbrace v_1, v_2, \ldots, v_r \rbrace$,
$$\beti{1}{\Gamma} = \beta_1 \vee \beta_2 \vee \ldots \vee \beta_r,$$ and
$\forall \: j \in [1, r]$, $\beta_{j} \not\in \C{\mid\!\sim}{\Gamma}$, $\MM{\Gamma}{\C{\mid\!\sim}{\Gamma}} \subseteq \M{\beta_{j}}$,
and $v_j \not\in \M{\neg\beta_{j}}$.
\\
As $\Mi{1}{\Gamma} \not= \emptyset$ and $\Mi{1}{\Gamma}$ is finite
(thanks to $(A1)$), such an element exists.
\\ \\
Suppose $i \in \mathbb{N}$, $i \geq 2$, and $\Mi{i}{\Gamma} \not= \emptyset$.
\\
Then, we denote by $\beti{i}{\Gamma}$ an element of $\cal F$, chosen arbitrarily, such that
\\
$\exists \: r \in \mathbb{N}^+$,
$\exists \: v_1, v_2, \ldots , v_r \in {\cal V}$, and
$\exists \: \beta_1, \beta_2, \ldots, \beta_r \in {\cal F}$ with
$\Mi{i}{\Gamma} = \lbrace v_1, v_2, \ldots, v_r \rbrace$,
$$\beti{i}{\Gamma} = \beta_1 \vee \beta_2 \vee \ldots \vee \beta_r,$$ and
$\forall \: j \in [1, r]$,
$\beta_{j} \not\in \C{\mid\!\sim}{\Gamma}$,
$\MM{\Gamma}{\C{\mid\!\sim}{\Gamma}} \setminus \Mi{1}{\Gamma} \cup \ldots \cup \Mi{i-1}{\Gamma} \subseteq \M{\beta_{j}}$,
and $v_j \not\in \M{\neg\beta_{j}}$.
\\
As $\Mi{i}{\Gamma} \not= \emptyset$ and $\Mi{i}{\Gamma}$ is finite, such an element exists.
\\ \\
Suppose $\Mp{\Gamma} \not= \emptyset$. Then,
$$
\bet{\Gamma} := \beti{1}{\Gamma} \vee \beti{2}{\Gamma} \vee \ldots \vee \beti{\n{\Gamma}}{\Gamma}
$$
As $\Mp{\Gamma} \not= \emptyset$, $\n{\Gamma} \geq 1$.
In~addition, we will show in Lemma~\ref{PREFsensfacile} below that $\n{\Gamma}$ is finite and 
$\forall \: i \in \mathbb{N}^+$ with $i \leq \n{\Gamma}$, $\Mi{i}{\Gamma} \not= \emptyset$.
Thus, $\bet{\Gamma}$ is well-defined.
\\
$$
\F{\Gamma} := \left \{ \begin{array}{ll} \lbrace \neg\bet{\Gamma} \rbrace & \textrm{if}\;  \Mp{\Gamma} \not= \emptyset \\
\emptyset & \textrm{otherwise} \end{array} \right .
$$
$$
\G{\Gamma} := \lbrace \alpha \in {\cal F} :
\alpha \not\in \C{\mid\!\sim}{\Gamma} ,\; \neg\alpha \not\in \C{\mid\!\sim}{\Gamma},
\;\textrm{and}\; \Td{\MMM{\Gamma}{\C{\mid\!\sim}{\Gamma}}{\alpha}}
\subseteq \C{\mid\!\sim}{\Gamma}\rbrace
$$
\end{definition}
Here are some quick results about the purely technical tools defined just above:

\begin{lemma} \label{PREFsensfacile}
Suppose $\cal L$ is a language, $\neg$ a unary connective of $\cal L$,
$\vee$ a binary connective of $\cal L$,
$\cal F$ the set of all wffs of $\cal L$,
$\langle {\cal F}, {\cal V}, \models \rangle$ a semantic structure satisfying $(A1)$, $\mid\!\sim$ a relation on
${\cal P}({\cal F}) \times {\cal F}$, $\Gamma \subseteq {\cal F}$, and $i, j \in \mathbb{N}^+$. Then,
\begin{description}
\item[$(0)$] if $i \not= j$, then $\Mi{i}{\Gamma} \cap \Mi{j}{\Gamma} = \emptyset$;
\item[$(1)$] if $\Mi{i}{\Gamma} = \emptyset$, then $\Mi{i+1}{\Gamma} = \emptyset$;
\item[$(2)$] $\Td{\MM{\Gamma}{\C{\mid\!\sim}{\Gamma}}} \subseteq \C{\mid\!\sim}{\Gamma}$ iff $\Mi{1}{\Gamma} = \emptyset$;
\item[$(3)$] if $i \geq 2$, then $\Td{\MM{\Gamma}{\C{\mid\!\sim}{\Gamma}} \setminus \Mi{1}{\Gamma} \cup \ldots \cup
\Mi{i-1}{\Gamma}} \subseteq \C{\mid\!\sim}{\Gamma}$ iff $\Mi{i}{\Gamma} = \emptyset$;
\item[$(4)$] $\n{\Gamma}$ is finite;
\item[$(5)$] if $i \leq \n{\Gamma}$, then $\Mi{i}{\Gamma} \not= \emptyset$;
\item[$(6)$] if $i > \n{\Gamma}$, then $\Mi{i}{\Gamma} = \emptyset$;
\item[$(7)$] if $\Mp{\Gamma} \not= \emptyset$, then $\Mp{\Gamma} = \Mi{1}{\Gamma} \cup \ldots \cup \Mi{\n{\Gamma}}{\Gamma}$;
\item[$(8)$] $\Td{\MM{\Gamma}{\C{\mid\!\sim}{\Gamma}} \setminus \Mp{\Gamma}} \subseteq \C{\mid\!\sim}{\Gamma}$.
\end{description}
\end{lemma}
\begin{proof} {\it Proofs of $(0)$, $(1)$, $(2)$, and $(3)$.} Trivial.
\\ \\
{\it Proof of $(4)$.} Obvious by $(0)$ and $(A1)$.
\\ \\
{\it Proof of $(5)$.} Suppose $\exists \: i \in \mathbb{N}^+$,
$\Mi{i}{\Gamma} = \emptyset$ and $i \leq \n{\Gamma}$.
\\
Then, by $(1)$, $\forall \: j \in \mathbb{N}^+$, $j \geq i$, $\Mi{j}{\Gamma} = \emptyset$.
\\
Thus, $|\lbrace j \in \mathbb{N}^+ : \Mi{j}{\Gamma} \not= \emptyset \rbrace| \leq i - 1 < \n{\Gamma}$, which is impossible.
\\ \\
{\it Proof of $(6)$.} Suppose $\exists \: i \in \mathbb{N}^+$, $\Mi{i}{\Gamma} \not= \emptyset$ and $i > \n{\Gamma}$.
\\
Then, by $(1)$, $\forall \: j \in \mathbb{N}^+$, $j \leq i$, $\Mi{j}{\Gamma} \not= \emptyset$.
\\
Thus, $|\lbrace j \in \mathbb{N}^+ : \Mi{j}{\Gamma} \not= \emptyset \rbrace| \geq i > \n{\Gamma}$, which is impossible.
\\ \\
{\it Proof of $(7)$.} Obvious by $(6)$.
\\ \\
{\it Proof of $(8)$.} Case~1: $\Mp{\Gamma} = \emptyset$.
\\
Then, $\Td{\MM{\Gamma}{\C{\mid\!\sim}{\Gamma}} \setminus \Mp{\Gamma}} =
\Td{\MM{\Gamma}{\C{\mid\!\sim}{\Gamma}}}$.
In addition, $\Mi{1}{\Gamma} = \emptyset$. Thus, by $(2)$, we are done.
\\
Case~2: $\Mp{\Gamma} \not= \emptyset$.
\\
Then, by $(7)$, $\Td{\MM{\Gamma}{\C{\mid\!\sim}{\Gamma}} \setminus \Mp{\Gamma}} = \Td{\MM{\Gamma}{\C{\mid\!\sim}{\Gamma}} \setminus \Mi{1}{\Gamma} \cup \ldots \cup
 \Mi{\n{\Gamma}}{\Gamma}}$.
 \\
In addition, $\n{\Gamma} + 1 \geq 2$ and, by $(6)$,
 $\Mi{\n{\Gamma} +1}{\Gamma} = \emptyset$. Thus, by $(3)$, we are done.\qed
\end{proof}
We turn to an important lemma. Its main goal is to show that the conditions
$(\mid\!\sim$$6)$, $(\mid\!\sim$$7)$, and $(\mid\!\sim$$8)$ are sufficient to establish
the following important equality: $\C{\mid\!\sim}{\Gamma} = \Td{\MMM{\Gamma}{\C{\mid\!\sim}{\Gamma}}{\monH{\Gamma}}}$, which provides a semantic definition of $\mid\!\sim$
(in the discriminative manner).

\begin{lemma}\label{PREF3sim}
Suppose $\cal L$ is a language, $\neg$ a unary connective of $\cal L$,
$\vee$ and $\wedge$ binary connectives of $\cal L$,
$\cal F$ the set of all wffs of $\cal L$,
$\langle {\cal F}, {\cal V}, \models \rangle$ a semantic structure satisfying $(A1)$ and $(A3)$, $\mid\!\sim$ a relation on ${\cal P}({\cal F}) \times {\cal F}$ satisfying $(\mid\!\sim$$6)$, $(\mid\!\sim$$7)$, and $(\mid\!\sim$$8)$,
and $\Gamma \subseteq {\cal F}$. Then,
\begin{description}
\item[$(0)$] if $\Mp{\Gamma} \not= \emptyset$, then
$\bet{\Gamma} \not\in \C{\mid\!\sim}{\Gamma}$;
\item[$(1)$] if $\Mp{\Gamma} \not= \emptyset$, then
$\MM{\Gamma}{\C{\mid\!\sim}{\Gamma}} \subseteq \M{\bet{\Gamma}}$;
\item[$(2)$] if $\Mp{\Gamma} \not= \emptyset$, then
$\Mp{\Gamma} \cap \M{\neg\bet{\Gamma}} = \emptyset$;
\item[$(3)$] if $\Mp{\Gamma} \not= \emptyset$, then
$\MM{\Gamma}{\C{\mid\!\sim}{\Gamma}} \setminus \Mp{\Gamma} \subseteq \M{\neg\bet{\Gamma}}$;
\item[$(4)$]
$\MM{\Gamma}{\C{\mid\!\sim}{\Gamma}} \setminus \Mp{\Gamma} =
\MMM{\Gamma}{\C{\mid\!\sim}{\Gamma}}{\F{\Gamma}}$;
\item[$(5)$] $\C{\mid\!\sim}{\Gamma} = \Td{\MMM{\Gamma}{\C{\mid\!\sim}{\Gamma}}{\F{\Gamma}}}$;
\item[$(6)$] $\MMM{\Gamma}{\C{\mid\!\sim}{\Gamma}}{\monH{\Gamma}} =
\MMM{\Gamma}{\C{\mid\!\sim}{\Gamma}}{\F{\Gamma}}$;
\item[$(7)$] $\C{\mid\!\sim}{\Gamma} = \Td{\MMM{\Gamma}{\C{\mid\!\sim}{\Gamma}}{\monH{\Gamma}}}$.
\end{description}
\end{lemma}

\begin{proof}
{\it Proof of $(0)$, $(1)$, and $(2)$.} Suppose $\Mp{\Gamma} \not= \emptyset$.
\\
Then, it suffices to show by induction: $\forall \: i \in [1, \n{\Gamma}]$,
\\
$p_3(i)$\quad $(\Mi{1}{\Gamma} \cup \ldots \cup \Mi{i}{\Gamma}) \cap
\M{\neg(\beti{1}{\Gamma} \vee \ldots \vee \beti{i}{\Gamma})} = \emptyset$;
\\
$p_2(i)$\quad $\MM{\Gamma}{\C{\mid\!\sim}{\Gamma}} \subseteq \M{\beti{1}{\Gamma} \vee \ldots \vee \beti{i}{\Gamma}}$;
\\
$p_1(i)$\quad $\beti{1}{\Gamma} \vee \ldots \vee \beti{i}{\Gamma} \not\in \C{\mid\!\sim}{\Gamma}$.
\\
As $\Mi{1}{\Gamma} \not= \emptyset$, $\exists \: r \in \mathbb{N}^+$,
$\exists \: v_1, v_2, \ldots , v_r \in {\cal V}$, and
$\exists \beta_1, \beta_2, \ldots, \beta_r \in {\cal F}$,
$\Mi{1}{\Gamma} = \lbrace v_1, \ldots , v_r \rbrace$,
\\
$\beti{1}{\Gamma} = \beta_1 \vee \ldots \vee \beta_r$, and
$\forall \: j \in [1, r]$, $\beta_{j} \not\in \C{\mid\!\sim}{\Gamma}$,
$\MM{\Gamma}{\C{\mid\!\sim}{\Gamma}} \subseteq \M{\beta_{j}}$,
and $v_j \not\in \M{\neg\beta_{j}}$.
\\
Then, it can be shown that:
\\
$(0.0)$\quad $p_3(1)$ holds;
\\
$(0.1)$\quad $p_2(1)$ holds;
\\
$(0.2)$\quad $p_1(1)$ holds.
\\
Now, let $i \in [1, \n{\Gamma} - 1]$ and suppose $p_1(i)$, $p_2(i)$, and $p_3(i)$ hold.
\\
As $\Mi{i+1}{\Gamma} \not= \emptyset$, $\exists \: r \in \mathbb{N}^+$,
$\exists \: v_1, v_2, \ldots , v_r \in {\cal V}$, and
$\exists \: \beta_1, \beta_2, \ldots, \beta_r \in {\cal F}$,
\\
$\Mi{i+1}{\Gamma} = \lbrace v_1, \ldots, v_r \rbrace$,
$\beti{i+1}{\Gamma} = \beta_1 \vee \ldots \vee \beta_r$, and
\\
$\forall \: j \in [1, r]$,
$\beta_{j} \not\in \C{\mid\!\sim}{\Gamma}$,
$\MM{\Gamma}{\C{\mid\!\sim}{\Gamma}} \setminus \Mi{1}{\Gamma} \cup \ldots \cup \Mi{i}{\Gamma} \subseteq \M{\beta_{j}}$,
and $v_j \not\in \M{\neg\beta_{j}}$.
\\
Then, it can be shown that:
\\
$(0.3)$\quad $p_3(i+1)$ holds;
\\
$(0.4)$\quad $p_2(i+1)$ holds.
\\
Before turning to $p_1(i+1)$, we need the following:
\\
$(0.5)$\quad $\beti{1}{\Gamma} \vee \ldots \vee \beti{i}{\Gamma} \vee \beta_1 \vee \beta_2 \vee \ldots \vee \beta_r
\not\in \C{\mid\!\sim}{\Gamma}$;
\\
$(0.6)$\quad $p_1(i+1)$ holds.
\\ \\
{\it Proof of $(0.0)$.} If $v_j \in \Mi{1}{\Gamma}$, then
$v_j \not\in \M{\neg\beta_{j}}$. But, by $(A3)$,
$\M{\neg\beti{1}{\Gamma}} \subseteq \M{\neg\beta_{j}}$.
\\ \\
{\it Proof of $(0.1)$.} We have $\MM{\Gamma}{\C{\mid\!\sim}{\Gamma}} \subseteq \M{\beta_1}$
which is, by $(A3)$, a subset of $\M{\beti{1}{\Gamma}}$.
\\ \\
{\it Proof of $(0.2)$.} It suffices to show by induction: $\forall \: j \in [1, r]$,
\\
$q(j)$\quad $\beta_1 \vee \ldots \vee \beta_j \not\in \C{\mid\!\sim}{\Gamma}$.
\\
Obviously, $q(1)$ holds.
\\
Let $j \in  [1, r-1]$. Suppose $q(j)$. We show $q(j+1)$.
\\
By $(A3)$, we have $\MM{\Gamma}{\C{\mid\!\sim}{\Gamma}} \subseteq \M{\beta_1 \vee \ldots \vee \beta_j}$.
\\
On the other hand, $\MMM{\Gamma}{\C{\mid\!\sim}{\Gamma}}{\neg(\beta_1 \vee \ldots \vee \beta_j)} \subseteq
\MM{\Gamma}{\C{\mid\!\sim}{\Gamma}} \subseteq
\M{\beta_{j+1}}$.
\\
Thus, by $q(j)$ and $(\mid\!\sim$$7)$
(where $\alpha$ is $\beta_1 \vee \ldots \vee \beta_j$ and $\beta$ is $\beta_{j+1}$), we get
$\beta_1 \vee \ldots \vee \beta_{j+1} \not\in \C{\mid\!\sim}{\Gamma}$.
\\ \\
{\it Proof of $(0.3)$}. Let $v \in \Mi{1}{\Gamma} \cup \ldots \cup \Mi{i+1}{\Gamma}$.
We show $v \not\in \M{\neg(\beti{1}{\Gamma} \vee \ldots \vee \beti{i+1}{\Gamma})}$.
\\
Case~1: $v \in \Mi{1}{\Gamma} \cup \ldots \cup \Mi{i}{\Gamma}$.
\\
Then, by $p_3(i)$, we have $v \not\in \M{\neg(\beti{1}{\Gamma} \vee \ldots \vee \beti{i}{\Gamma})}$.
But, by $(A3)$, $\M{\neg(\beti{1}{\Gamma} \vee \ldots \vee \beti{i+1}{\Gamma})} \subseteq
\M{\neg(\beti{1}{\Gamma} \vee \ldots \vee \beti{i}{\Gamma})}$.
\\
Case~2: $v \in \Mi{i+1}{\Gamma}$.
\\
Then, $\exists \: j \in [1, r]$, $v = v_j$.
Thus, $v \not\in \M{\neg\beta_{j}}$.
But, by $(A3)$, $\M{\neg(\beti{1}{\Gamma} \vee \ldots \vee \beti{i+1}{\Gamma})}
\subseteq \M{\neg\beti{i+1}{\Gamma}} \subseteq \M{\neg\beta_{j}}$.
\\ \\
{\it Proof of $(0.4)$}. By $p_2(i)$,
$\MM{\Gamma}{\C{\mid\!\sim}{\Gamma}} \subseteq \M{\beti{1}{\Gamma} \vee \ldots \vee \beti{i}{\Gamma}}$
which is, by $(A3)$, a subset of $\M{\beti{1}{\Gamma} \vee \ldots \vee \beti{i+1}{\Gamma}}$.
\\ \\
{\it Proof of $(0.5)$}. It suffices to show by induction $\forall \: j \in [1, r]$:
\\
$q(j)$\quad $\beti{1}{\Gamma} \vee \ldots \vee \beti{i}{\Gamma} \vee \beta_1 \vee \ldots \vee \beta_j \not\in \C{\mid\!\sim}{\Gamma}$.
\\
We will show:
\\
$(0.5.0)$\quad $\MMM{\Gamma}{\C{\mid\!\sim}{\Gamma}}{\neg(\beti{1}{\Gamma} \vee \ldots \vee \beti{i}{\Gamma})}
\subseteq \M{\beta_1}$.
\\
Then, by $p_1(i)$, $p_2(i)$, $(0.5.0)$, and $(\mid\!\sim$$7)$
(where $\alpha$ is $\beti{1}{\Gamma} \vee \ldots \vee \beti{i}{\Gamma}$ and $\beta$ is $\beta_1$), $q(1)$ holds.
\\
Now, let $j \in [1, r-1]$ and suppose $q(j)$.
\\
Then, we will show:
\\
$(0.5.1)$\quad $\MMM{\Gamma}{\C{\mid\!\sim}{\Gamma}}{\neg(\beti{1}{\Gamma} \vee \ldots \vee \beti{i}{\Gamma} \vee \beta_1 \vee \ldots \vee \beta_j)} \subseteq \M{\beta_{j +1}}$.
\\
In addition, by $p_2(i)$ and $(A3)$, we get:
\\
$(0.5.2)$\quad $\MM{\Gamma}{\C{\mid\!\sim}{\Gamma}} \subseteq
\M{\beti{1}{\Gamma} \vee \ldots \vee \beti{i}{\Gamma} \vee \beta_1 \vee \ldots \vee \beta_j}$.
\\
By, $(0.5.1)$, $(0.5.2)$, $q(j)$, and $(\mid\!\sim$$7)$
(where $\alpha$ is $\beti{1}{\Gamma} \vee \ldots \vee \beti{i}{\Gamma} \vee \beta_1 \vee \ldots \vee \beta_j$
and $\beta$ is $\beta_{j+1}$),
\\
we get that $q(j+1)$ holds.
\\ \\
{\it Proof of $(0.5.0)$.}
Let $v \in \MMM{\Gamma}{\C{\mid\!\sim}{\Gamma}}{\neg(\beti{1}{\Gamma} \vee \ldots \vee \beti{i}{\Gamma})}$. Then, $v \in \M{\neg(\beti{1}{\Gamma} \vee \ldots \vee \beti{i}{\Gamma})}$.
\\
Thus, by $p_3(i)$, $v \not\in \Mi{1}{\Gamma} \cup \ldots \cup \Mi{i}{\Gamma}$.
Therefore, $v \in \MM{\Gamma}{\C{\mid\!\sim}{\Gamma}} \setminus \Mi{1}{\Gamma} \cup \ldots \cup \Mi{i}{\Gamma} \subseteq \M{\beta_1}$.
\\ \\
{\it Proof of $(0.5.1)$.}
Let $v \in \MMM{\Gamma}{\C{\mid\!\sim}{\Gamma}}{\neg(\beti{1}{\Gamma} \vee \ldots \vee \beti{i}{\Gamma} \vee
\beta_1 \vee \ldots \vee \beta_j)}$.
Then, by $(A3)$, $v \in \M{\neg(\beti{1}{\Gamma} \vee \ldots \vee \beti{i}{\Gamma})}$.
\\
Therefore, by $p_3(i)$, $v \not\in \Mi{1}{\Gamma} \cup \ldots \cup \Mi{i}{\Gamma}$.
Therefore, $v \in \MM{\Gamma}{\C{\mid\!\sim}{\Gamma}} \setminus \Mi{1}{\Gamma} \cup \ldots \cup \Mi{i}{\Gamma} \subseteq \M{\beta_{j+1}}$.
\\ \\
{\it Proof of $(0.6)$.}
By $p_2(i)$ and $(A3)$, we get
$\MM{\Gamma}{\C{\mid\!\sim}{\Gamma}} \subseteq
\M{\beti{1}{\Gamma} \vee \ldots \vee \beti{i}{\Gamma}} \subseteq
\M{\beti{1}{\Gamma} \vee \ldots \vee \beti{i}{\Gamma} \vee \beta_1 \vee \ldots \vee \beta_r}$.
\\
In addition, by $(A3)$, we get
$\M{\neg(\beti{1}{\Gamma} \vee \ldots \vee \beti{i}{\Gamma} \vee \beta_1 \vee \ldots \vee \beta_r)}
= \M{\neg(\beti{1}{\Gamma} \vee \ldots \vee \beti{i+1}{\Gamma})}$.
\\
Therefore, by $(0.5)$ and $(\mid\!\sim$$6)$
(where $\alpha$ is $\beti{1}{\Gamma} \vee \ldots \vee \beti{i+1}{\Gamma}$
and $\beta$ is $\beti{1}{\Gamma} \vee \ldots \vee \beti{i}{\Gamma} \vee \beta_1 \vee \ldots \vee \beta_r$),
\\
we get that $p_1(i+1)$ holds.
\\ \\
{\it Proof of $(3)$.} Suppose $\Mp{\Gamma} \not= \emptyset$,
$v \in \MM{\Gamma}{\C{\mid\!\sim}{\Gamma}} \setminus \Mp{\Gamma}$,
and $v \not\in \M{\neg\bet{\Gamma}}$.
\\
Then, by $(0)$, $(1)$, and definition of $\Mi{i}{\Gamma}$, we get $v \in \Mi{\n{\Gamma}+1}{\Gamma}$, which is impossible by Lemma~\ref{PREFsensfacile}~$(6)$.
\\ \\
{\it Proof of $(4)$.} Case~1: $\Mp{\Gamma} \not= \emptyset$.
\\
By $(3)$, we get one direction: $\MM{\Gamma}{\C{\mid\!\sim}{\Gamma}} \setminus \Mp{\Gamma}
\subseteq \MMM{\Gamma}{\C{\mid\!\sim}{\Gamma}}{\neg\bet{\Gamma}}$.
\\
By $(2)$, we get the other direction:
$\MMM{\Gamma}{\C{\mid\!\sim}{\Gamma}}{\neg\bet{\Gamma}} \subseteq
\MM{\Gamma}{\C{\mid\!\sim}{\Gamma}} \setminus \Mp{\Gamma}$.
\\
Case~2: $\Mp{\Gamma} = \emptyset$.
\\
Then, obviously, $\MM{\Gamma}{\C{\mid\!\sim}{\Gamma}} \setminus \Mp{\Gamma} =
\MM{\Gamma}{\C{\mid\!\sim}{\Gamma}} =
\MMM{\Gamma}{\C{\mid\!\sim}{\Gamma}}{\F{\Gamma}}$.
\\ \\
{\it Proof of $(5)$}. Direction: ``$\subseteq$''.
\\
Case~1: $\Mp{\Gamma} \not= \emptyset$.
\\
Suppose the contrary of what we want to show, i.e. suppose
$\exists \: \alpha \in \C{\mid\!\sim}{\Gamma}$, $\alpha \not\in \Td{\MMM{\Gamma}{\C{\mid\!\sim}{\Gamma}}{\neg\bet{\Gamma}}}$.
\\
Then, $\MMM{\Gamma}{\C{\mid\!\sim}{\Gamma}}{\neg\bet{\Gamma}} \subseteq
\M{\C{\mid\!\sim}{\Gamma}} \subseteq \M{\alpha}$. Thus,
$\MMM{\Gamma}{\C{\mid\!\sim}{\Gamma}}{\neg\bet{\Gamma}} \subseteq \M{\neg\alpha}$.
\\
Consequently, by $(0)$, $(1)$, and $(\mid\!\sim$$6)$, we get $\alpha \not\in \C{\mid\!\sim}{\Gamma}$,
which is impossible.
\\
Case~2: $\Mp{\Gamma} = \emptyset$.
\\
Let $\alpha \in \C{\mid\!\sim}{\Gamma}$. Then,
$\MM{\Gamma}{\C{\mid\!\sim}{\Gamma}} \subseteq
\M{\C{\mid\!\sim}{\Gamma}} \subseteq \M{\alpha}$. In addition, by $(\mid\!\sim$$8)$,
$\MM{\Gamma}{\C{\mid\!\sim}{\Gamma}} \not\subseteq \M{\neg\alpha}$.
\\
Consequently, $\alpha \in \Td{\MM{\Gamma}{\C{\mid\!\sim}{\Gamma}}} = \Td{\MMM{\Gamma}{\C{\mid\!\sim}{\Gamma}}{\F{\Gamma}}}$.

Direction: ``$\supseteq$''.
Obvious by $(4)$ and Lemma~\ref{PREFsensfacile}~$(8)$.
\\ \\
{\it Proof of $(6)$.} Direction:  ``$\subseteq$''.
\\
Case~1: $\Mp{\Gamma} = \emptyset$.
\\
Case~1.1: $\monHi{1}{\Gamma} \not= \emptyset$.
\\
Then, $\exists \: \alpha \in {\cal F}$, $\alpha \not\in \C{\mid\!\sim}{\Gamma}$,
$\MM{\Gamma}{\C{\mid\!\sim}{\Gamma}}
\subseteq \M{\alpha}$, and $\MM{\Gamma}{\C{\mid\!\sim}{\Gamma}} \not\subseteq \M{\neg\alpha}$.
Thus, $\alpha \in \Td{\MM{\Gamma}{\C{\mid\!\sim}{\Gamma}}}$.
Therefore, by $(5)$, $\alpha \in \C{\mid\!\sim}{\Gamma}$, which is impossible.
\\
Case~1.2: $\monHi{1}{\Gamma} = \emptyset$.
\\
Clearly, $\forall \: i \in \mathbb{N}^+$,  if $\monHi{i}{\Gamma} = \emptyset$, then $\monHi{i+1}{\Gamma} = \emptyset$. Therefore, $\monH{\Gamma} = \emptyset = \F{\Gamma}$.
\\
Case~2: $\Mp{\Gamma} \not= \emptyset$.
\\
As, $\Mp{\Gamma} \subseteq \MM{\Gamma}{\C{\mid\!\sim}{\Gamma}}$, we get, by $(2)$,
$\MM{\Gamma}{\C{\mid\!\sim}{\Gamma}} \not\subseteq \M{\neg\bet{\Gamma}}$.
\\
Thus, by $(0)$ and $(1)$, we get $\neg\bet{\Gamma} \in \monHi{1}{\Gamma} \subseteq \monH{\Gamma}$.
Therefore, $\M{\monH{\Gamma}} \subseteq \M{\F{\Gamma}}$.

Direction:  ``$\supseteq$''.
\\
Case~1: $\Mp{\Gamma} = \emptyset$.
\\
Verbatim the proof of Case~1 of direction~``$\subseteq$''.
\\
Case~2: $\Mp{\Gamma} \not= \emptyset$.
\\
Then, the following holds:
\\
$(6.0)$\quad $\forall i \in \mathbb{N}^+$, $\MMM{\Gamma}{\C{\mid\!\sim}{\Gamma}}{\neg\bet{\Gamma}} \subseteq \MMM{\Gamma}{\C{\mid\!\sim}{\Gamma}}{\monHi{1}{\Gamma}, \ldots, \monHi{i}{\Gamma}}$.
\\
Now, suppose the contrary of what we want to show,
i.e. suppose
\\
$\exists \: v \in \MMM{\Gamma}{\C{\mid\!\sim}{\Gamma}}{\neg\bet{\Gamma}}$,
$v \not\in \MMM{\Gamma}{\C{\mid\!\sim}{\Gamma}}{\monH{\Gamma}}$.
Then, $v \not\in \M{\monH{\Gamma}}$. But, clearly,
$\M{\monH{\Gamma}} = \bigcap_{i \in \mathbb{N}^+} \M{\monHi{i}{\Gamma}}$.
Therefore, $\exists \: i \in \mathbb{N}^+$, $v \not\in \M{\monHi{i}{\Gamma}}$,
which is impossible by $(6.0)$.
\\ \\
{\it Proof of $(6.0)$.} We show by induction: $\forall i \in \mathbb{N}^+$,
\\
$p(i)$\quad $\MMM{\Gamma}{\C{\mid\!\sim}{\Gamma}}{\neg\bet{\Gamma}} \subseteq \MMM{\Gamma}{\C{\mid\!\sim}{\Gamma}}{\monHi{1}{\Gamma}, \ldots, \monHi{i}{\Gamma}}$.
\\
We will show
\\
$(6.0.0)$\quad $p(1)$ holds.
\\
Let $i \in \mathbb{N}^+$, suppose $p(i)$ holds, and suppose $p(i+1)$ does not hold.
\\
Then, $\exists \: v \in \MMM{\Gamma}{\C{\mid\!\sim}{\Gamma}}{\neg\bet{\Gamma}}$,
$v \not\in \MMM{\Gamma}{\C{\mid\!\sim}{\Gamma}}{\monHi{1}{\Gamma}, \ldots, \monHi{i+1}{\Gamma}}$.
\\
Thus, $\exists \: j \in [1, i + 1]$, $v \not\in \M{\monHi{j}{\Gamma}}$.
\\
Case~1: $j = 1$.
\\
Then, $\exists \: \beta \in {\cal F}$,
$\MM{\Gamma}{\C{\mid\!\sim}{\Gamma}} \subseteq \M{\beta}$, $\beta \not\in \C{\mid\!\sim}{\Gamma}$, and $v \not\in \M{\neg\beta}$.
\\
Thus $v \in \Mi{1}{\Gamma} \cap \M{\neg\bet{\Gamma}}$, which is impossible by $(2)$.
\\
Case~2: $j \geq 2$.
\\
Then, $\exists \: \beta \in {\cal F}$,
$\MMM{\Gamma}{\C{\mid\!\sim}{\Gamma}}{\monHi{1}{\Gamma}, \ldots, \monHi{j-1}{\Gamma}} \subseteq \M{\beta}$,
$\beta \not\in \C{\mid\!\sim}{\Gamma}$, and $v \not\in \M{\neg\beta}$.
\\
But, by Lemma~\ref{PREFsensfacile}~$(7)$, by $(4)$, and $p(i)$, we get
\\
$\MM{\Gamma}{\C{\mid\!\sim}{\Gamma}} \setminus \Mi{1}{\Gamma} \cup \ldots \cup \Mi{\n{\Gamma}}{\Gamma} =
\MM{\Gamma}{\C{\mid\!\sim}{\Gamma}} \setminus \Mp{\Gamma} =
\MMM{\Gamma}{\C{\mid\!\sim}{\Gamma}}{\neg\bet{\Gamma}} \subseteq
\MMM{\Gamma}{\C{\mid\!\sim}{\Gamma}}{\monHi{1}{\Gamma}, \ldots, \monHi{i}{\Gamma}} \subseteq
\MMM{\Gamma}{\C{\mid\!\sim}{\Gamma}}{\monHi{1}{\Gamma}, \ldots, \monHi{j-1}{\Gamma}} \subseteq \M{\beta}$.
\\
Therefore, $v \in \Mi{\n{\Gamma} +1}{\Gamma}$, which is impossible by Lemma~\ref{PREFsensfacile}~$(6)$.
\\ \\
{\it Proof of $(6.0.0)$.} Suppose the contrary of what we want to show, i.e.
\\
suppose $\exists \: v \in \MMM{\Gamma}{\C{\mid\!\sim}{\Gamma}}{\neg\bet{\Gamma}}$, $v \not\in
\MMM{\Gamma}{\C{\mid\!\sim}{\Gamma}}{\monHi{1}{\Gamma}}$.
\\
Then, $v \not\in \M{\monHi{1}{\Gamma}}$. Thus, $\exists \: \beta \in {\cal F}$,
$\MM{\Gamma}{\C{\mid\!\sim}{\Gamma}} \subseteq \M{\beta}$, $\beta \not\in \C{\mid\!\sim}{\Gamma}$, and $v \not\in \M{\neg\beta}$.
\\
Thus $v \in \Mi{1}{\Gamma}$. Therefore, $v \in \Mp{\Gamma} \cap \M{\neg\bet{\Gamma}}$, which is impossible by $(2)$.
\\ \\
{\it Proof of $(7)$.} Obvious by $(5)$ and $(6)$.\qed
\end{proof}
We turn to a second important lemma. Its main purpose is to show that any DP choice function
$\mu$ representing (in the discriminative manner) a relation $\mid\!\sim$ satisfies the following
equality: $\mu(\M{\Gamma}) = \MMM{\Gamma}{\C{\mid\!\sim}{\Gamma}}{\monH{\Gamma}}$,
which enables us to define $\mu$ from $\mid\!\sim$.

\begin{lemma} \label{PREFPrf2ConArgSyn}
Suppose $\cal L$ is a language, $\neg$ a unary connective of $\cal L$,
$\vee$ and $\wedge$ binary connectives of $\cal L$,
$\cal F$ the set of all wffs of $\cal L$,
$\langle {\cal F}, {\cal V}, \models \rangle$ a semantic structure satisfying $(A1)$ and $(A3)$,
${\bf V} \subseteq {\cal P}({\cal V})$,
$\mu$ a DP choice function from $\bf D$ to $\bf V$, $\mid\!\sim$ the
relation on ${\cal P}({\cal F}) \times {\cal F}$ such that
$\forall \: \Gamma \subseteq {\cal F}$, $\C{\mid\!\sim}{\Gamma}
= \Td{\mu(\M{\Gamma})}$, and $\Gamma \subseteq {\cal F}$.
Then:
\begin{description}
\item[$(0)$] $\mu(\M{\Gamma}) \subseteq \MM{\Gamma}{\C{\mid\!\sim}{\Gamma}}$;
\item[$(1)$] $\mid\!\sim$ satisfies $(\mid\!\sim$$6)$;
\item[$(2)$] $\mid\!\sim$ satisfies $(\mid\!\sim$$7)$;
\item[$(3)$] $\mid\!\sim$ satisfies $(\mid\!\sim$$8)$;
\item[$(4)$] $\Mp{\Gamma} \cap \mu(\M{\Gamma}) = \emptyset$;
\item[$(5)$] $\MMM{\Gamma}{\C{\mid\!\sim}{\Gamma}}{\Tc{\mu(\M{\Gamma})}} = \mu(\M{\Gamma})$;
\item[$(6)$] if $\Mp{\Gamma} \not= \emptyset$, then
$\MMM{\Gamma}{\C{\mid\!\sim}{\Gamma}}{\monH{\Gamma}} = \mu(\M{\Gamma})$.
\end{description}
If $\langle {\cal F}, {\cal V}, \models \rangle$ satisfies $(A2)$ too, then:
\begin{description}
\item[$(7)$] if $\Mp{\Gamma} = \emptyset$, then $\M{\G{\Gamma}} = \M{\Tc{\mu(\M{\Gamma})}}$;
\item[$(8)$] if $\Mp{\Gamma} = \emptyset$, then $\MM{\Gamma}{\C{\mid\!\sim}{\Gamma}} \subseteq \M{\G{\Gamma}}$;
\item[$(9)$] $\MMM{\Gamma}{\C{\mid\!\sim}{\Gamma}}{\monH{\Gamma}} = \mu(\M{\Gamma})$.
\end{description}
If $\mu$ is coherency preserving, then again:
\begin{description}
\item[$(10)$] $\MMM{\Gamma}{\C{\mid\!\sim}{\Gamma}}{\monH{\Gamma}} = \mu(\M{\Gamma})$.
\end{description}
\end{lemma}

\begin{proof}
{\it Proof of $(0)$.} We show $\mu(\M{\Gamma}) \subseteq \M{\C{\mid\!\sim}{\Gamma}}$.
Let $v \in \mu(\M{\Gamma})$ and $\alpha \in \C{\mid\!\sim}{\Gamma}$.
\\
Then, $\alpha \in \Td{\mu(\M{\Gamma})}$. Thus, $\mu(\M{\Gamma}) \subseteq \M{\alpha}$.
Thus, $v \in \M{\alpha}$ and we are done.
\\
In addition, obviously, $\mu(\M{\Gamma}) \subseteq \M{\Gamma}$.
Therefore, $\mu(\M{\Gamma}) \subseteq \M{\Gamma} \cap \M{\C{\mid\!\sim}{\Gamma}}
= \MM{\Gamma}{\C{\mid\!\sim}{\Gamma}}$.
\\ \\
{\it Proof of $(1)$}.
Let $\alpha, \beta \in {\cal F}$ and suppose $\beta \in \CC{\vdash}{\Gamma}{\C{\mid\!\sim}{\Gamma}} \setminus \C{\mid\!\sim}{\Gamma}$ and $\neg\alpha \in \CCC{\vdash}{\Gamma}{\C{\mid\!\sim}{\Gamma}}{\neg\beta}$.
\\
Then, by $(0)$, $\mu(\M{\Gamma}) \subseteq \MM{\Gamma}{\C{\mid\!\sim}{\Gamma}} \subseteq \M{\beta}$.
But, $\beta \not\in \C{\mid\!\sim}{\Gamma} = \Td{\mu(\M{\Gamma})}$.
Thus, $\mu(\M{\Gamma}) \subseteq \M{\neg\beta}$.
\\
Consequently,
$\mu(\M{\Gamma}) \subseteq
\MMM{\Gamma}{\C{\mid\!\sim}{\Gamma}}{\neg\beta} \subseteq
\M{\neg\alpha}$. Therefore, $\alpha \not\in \Td{\mu(\M{\Gamma})} = \C{\mid\!\sim}{\Gamma}$.
\\ \\
{\it Proof of $(2)$.} Let $\alpha, \beta \in {\cal F}$ and suppose
$\alpha \in \CC{\vdash}{\Gamma}{\C{\mid\!\sim}{\Gamma}} \setminus \C{\mid\!\sim}{\Gamma}$ and
$\beta \in \CCC{\vdash}{\Gamma}{\C{\mid\!\sim}{\Gamma}}{\neg\alpha} \setminus \C{\mid\!\sim}{\Gamma}$.
\\
Then, by $(0)$, $\mu(\M{\Gamma}) \subseteq \MM{\Gamma}{\C{\mid\!\sim}{\Gamma}} \subseteq \M{\alpha}$.
But, $\alpha \not\in \Td{\mu(\M{\Gamma})}$. Thus, $\mu(\M{\Gamma}) \subseteq \M{\neg\alpha}$.
\\
Thus, $\mu(\M{\Gamma}) \subseteq \MMM{\Gamma}{\C{\mid\!\sim}{\Gamma}}{\neg\alpha}
\subseteq \M{\beta}$. But, $\beta \not\in \Td{\mu(\M{\Gamma})}$. Therefore
$\mu(\M{\Gamma}) \subseteq \M{\neg\beta}$.
\\
Thus, by $(A3)$, $\mu(\M{\Gamma}) \subseteq \M{\neg\alpha} \cap \M{\neg\beta} =
\M{\neg(\alpha \vee \beta)}$. Consequently, $\alpha \vee \beta \not\in \Td{\mu(\M{\Gamma})} = \C{\mid\!\sim}{\Gamma}$.
\\ \\
{\it Proof of $(3)$.} Let $\alpha \in \C{\mid\!\sim}{\Gamma}$.
Then, $\alpha \in \Td{\mu(\M{\Gamma})}$. Thus,
$\mu(\M{\Gamma}) \not\subseteq \M{\neg\alpha}$.
\\
Thus, by $(0)$, $\MM{\Gamma}{\C{\mid\!\sim}{\Gamma}} \not\subseteq \M{\neg\alpha}$.
\\ \\
{\it Proof of $(4)$.} Case~1: $\Mp{\Gamma} = \emptyset$. Obvious.
\\
Case~2: $\Mp{\Gamma} \not= \emptyset$.
\\
It is sufficient to show by induction: $\forall \: i \in [1, \n{\Gamma}]$,
\\
$p(i)$\quad $(\Mi{1}{\Gamma} \cup \ldots \cup \Mi{i}{\Gamma}) \cap \mu(\M{\Gamma}) = \emptyset$.
\\
We will show:
\\
$(4.0)$\quad $p(1)$ holds.
\\
Let $i \in [1, \n{\Gamma}-1]$. Suppose $p(i)$. We show $p(i+1)$.
\\
Case~1: $\Mi{i+1}{\Gamma} \cap \mu(\M{\Gamma}) = \emptyset$.
\\
Then, by $p(i)$, we obviously get $p(i+1)$.
\\
Case~2: $\exists \: v \in \Mi{i+1}{\Gamma} \cap \mu(\M{\Gamma})$.
\\
Then, $\exists \: \beta \in {\cal F}$, $\beta \not\in \C{\mid\!\sim}{\Gamma}$,
$\MM{\Gamma}{\C{\mid\!\sim}{\Gamma}} \setminus \Mi{1}{\Gamma} \cup \ldots \cup \Mi{i}{\Gamma} \subseteq \M{\beta}$, and $v \not\in \M{\neg\beta}$.
\\
Therefore, by $(0)$ and $p(i)$, $\mu(\M{\Gamma}) \subseteq \MM{\Gamma}{\C{\mid\!\sim}{\Gamma}}
\setminus \Mi{1}{\Gamma} \cup \ldots \cup \Mi{i}{\Gamma} \subseteq \M{\beta}$.
But, $\mu(\M{\Gamma}) \not\subseteq \M{\neg\beta}$.
\\
Consequently, $\beta \in \Td{\mu(\M{\Gamma})} = \C{\mid\!\sim}{\Gamma}$, which is impossible.
\\ \\
{\it Proof of $(4.0)$.} Suppose the contrary of $p(1)$, i.e. suppose
$\exists \: v \in \Mi{1}{\Gamma} \cap \mu(\M{\Gamma})$.
\\
Then, $\exists \: \beta \in {\cal F}$, $\beta \not\in \C{\mid\!\sim}{\Gamma}$,
$\MM{\Gamma}{\C{\mid\!\sim}{\Gamma}} \subseteq \M{\beta}$ and $v \not\in \M{\neg\beta}$.
\\
Therefore, by $(0)$, $\mu(\M{\Gamma}) \subseteq \M{\beta}$.
On the other hand, $\mu(\M{\Gamma}) \not\subseteq \M{\neg\beta}$.
\\
Therefore, $\beta \in \Td{\mu(\M{\Gamma})} = \C{\mid\!\sim}{\Gamma}$, which is impossible.
\\ \\
{\it Proof of $(5)$.} As $\mu(\M{\Gamma}) \in {\bf D}$,
$\exists \: \Gamma' \subseteq {\cal F}$, $\M{\Gamma'} = \mu(\M{\Gamma})$.
\\
Therefore, $\M{\T{\mu(\M{\Gamma})}} = \M{\T{\M{\Gamma'}}} =
\M{\Gamma'} = \mu(\M{\Gamma})$.
\\
Thus,
$\MMM{\Gamma}{\C{\mid\!\sim}{\Gamma}}{\Tc{\mu(\M{\Gamma})}}
= \MMM{\Gamma}{\Td{\mu(\M{\Gamma})}}{\Tc{\mu(\M{\Gamma})}}
= \MM{\Gamma}{\T{\mu(\M{\Gamma})}}$.
But, $\Gamma \subseteq \T{\mu(\M{\Gamma})}$.
\\
Therefore,
$\MM{\Gamma}{\T{\mu(\M{\Gamma})}} = \M{\T{\mu(\M{\Gamma})}}
= \mu(\M{\Gamma})$.
\\ \\
{\it Proof of $(6)$.} Suppose $\Mp{\Gamma} \not= \emptyset$. Direction:~``$\subseteq$''.
\\
Case~1: $\exists \: v \in \MM{\Gamma}{\C{\mid\!\sim}{\Gamma}} \setminus \Mi{1}{\Gamma} \cup \ldots \cup \Mi{\n{\Gamma}}{\Gamma}$,
$v \not\in \M{\Tc{\mu(\M{\Gamma})}}$.
\\
Then, $\exists \: \alpha \in \Tc{\mu(\M{\Gamma})}$, $v \not\in \M{\alpha}$.
\\
By Lemma~\ref{PREF3sim}~$(3)$, Lemma~\ref{PREFsensfacile}~$(7)$, and $(A3)$, $\MM{\Gamma}{\C{\mid\!\sim}{\Gamma}} \setminus \Mi{1}{\Gamma}
\cup \ldots \cup \Mi{\n{\Gamma}}{\Gamma} \subseteq
\M{\neg\bet{\Gamma}} \subseteq \M{\neg(\bet{\Gamma} \wedge \alpha)}$.
\\
By $(0)$ and Lemma~\ref{PREF3sim}~$(1)$,
$\mu(\M{\Gamma}) \subseteq \M{\bet{\Gamma}} \cap \M{\alpha}
= \M{\neg\neg(\bet{\Gamma} \wedge \alpha)}$.
\\
Therefore, $\neg(\bet{\Gamma} \wedge \alpha) \not\in \Td{\mu(\M{\Gamma})} = \C{\mid\!\sim}{\Gamma}$.
\\
In addition, $v \not\in \M{\alpha} \supseteq \M{\neg\neg(\bet{\Gamma} \wedge \alpha)}$.
\\
Consequently, $v \in \Mi{\n{\Gamma}+1}{\Gamma}$
(take $\neg(\bet{\Gamma} \wedge \alpha)$
for the $\beta$ of the definition of $\Mi{i}{\Gamma}$).
\\
Therefore, by Lemma~\ref{PREFsensfacile}~$(6)$, we get a contradiction.
\\
Case~2: $\MM{\Gamma}{\C{\mid\!\sim}{\Gamma}} \setminus \Mi{1}{\Gamma} \cup \ldots \cup \Mi{\n{\Gamma}}{\Gamma}
\subseteq \M{\Tc{\mu(\M{\Gamma})}}$.
\\
Then, by Lemma~\ref{PREF3sim}~$(6)$, Lemma~\ref{PREF3sim}~$(4)$,
Lemma~\ref{PREFsensfacile}~$(7)$, and by $(5)$, we get
\\
$\MMM{\Gamma}{\C{\mid\!\sim}{\Gamma}}{\monH{\Gamma}} =
\MM{\Gamma}{\C{\mid\!\sim}{\Gamma}} \setminus \Mi{1}{\Gamma} \cup \ldots \cup \Mi{\n{\Gamma}}{\Gamma}
\subseteq \MMM{\Gamma}{\C{\mid\!\sim}{\Gamma}}{\Tc{\mu(\M{\Gamma})}} = \mu(\M{\Gamma})$.

Direction: ``$\supseteq$''.
\\
By $(0)$, $(4)$, Lemma~\ref{PREF3sim}~$(4)$, and Lemma~\ref{PREF3sim}~$(6)$, we get
\\
$\mu(\M{\Gamma}) \subseteq \MM{\Gamma}{\C{\mid\!\sim}{\Gamma}} \setminus \Mp{\Gamma} = \MMM{\Gamma}{\C{\mid\!\sim}{\Gamma}}{\F{\Gamma}} =
\MMM{\Gamma}{\C{\mid\!\sim}{\Gamma}}{\monH{\Gamma}}$.
\\ \\
{\it Proof of $(7)$.} Suppose $\Mp{\Gamma} = \emptyset$.
Direction: ``$\supseteq$''.
\\
Suppose the contrary of what we want to show, i.e. suppose
$\exists \: v \in \M{\Tc{\mu(\M{\Gamma})}}$, $v \not\in \M{\G{\Gamma}}$.
\\
Then, $\exists \: \alpha \in \G{\Gamma}$, $v \not\in \M{\alpha}$.
\\
Case~1 : $\alpha \in \T{\MM{\Gamma}{\C{\mid\!\sim}{\Gamma}}}$.
\\
As $\alpha \in \G{\Gamma}$, $\alpha \not\in \C{\mid\!\sim}{\Gamma}$.
Thus, by Lemma~\ref{PREF3sim}~$(5)$,
$\alpha \not\in \Td{\MM{\Gamma}{\C{\mid\!\sim}{\Gamma}}}$.
\\
Therefore,
$\alpha \in \Tc{\MM{\Gamma}{\C{\mid\!\sim}{\Gamma}}}$.
Consequently, by $(0)$, $\alpha \in \Tc{\mu(\M{\Gamma})}$.
\\
Thus, $v \in \M{\alpha}$, which is impossible.
\\
Case~2: $\neg\alpha \in \T{\MM{\Gamma}{\C{\mid\!\sim}{\Gamma}}}$.
\\
As $\alpha \in \G{\Gamma}$, $\neg\alpha \not\in \C{\mid\!\sim}{\Gamma}$.
Thus, by Lemma~\ref{PREF3sim}~$(5)$,
$\neg\alpha \not\in \Td{\MM{\Gamma}{\C{\mid\!\sim}{\Gamma}}}$.
\\
Therefore,
$\neg\alpha \in \Tc{\MM{\Gamma}{\C{\mid\!\sim}{\Gamma}}}$.
Consequently, by $(A3)$, $\alpha \in \Tc{\MM{\Gamma}{\C{\mid\!\sim}{\Gamma}}}$.
\\
Therefore, by $(0)$, $\alpha \in \Tc{\mu(\M{\Gamma})}$.
Thus, $v \in \M{\alpha}$, which is impossible.
\\
Case~3 : $\alpha \not\in \T{\MM{\Gamma}{\C{\mid\!\sim}{\Gamma}}}$ and
$\neg\alpha \not\in \T{\MM{\Gamma}{\C{\mid\!\sim}{\Gamma}}}$.
\\
Then, by $(A2)$, $\MMM{\Gamma}{\C{\mid\!\sim}{\Gamma}}{\alpha} \not\subseteq \M{\neg\alpha}$.
Therefore, $\alpha \in \Td{\MMM{\Gamma}{\C{\mid\!\sim}{\Gamma}}{\alpha}}$.
\\
But, $\alpha \in \G{\Gamma}$. Thus, $\Td{\MMM{\Gamma}{\C{\mid\!\sim}{\Gamma}}{\alpha}} \subseteq
\C{\mid\!\sim}{\Gamma}$. Thus, $\alpha \in \C{\mid\!\sim}{\Gamma}$.
Thus, $\alpha \not\in \G{\Gamma}$, impossible.

Direction: ``$\subseteq$''.
\\
Suppose the contrary of what we want to show, i.e. suppose
$\exists \: v \in \M{\G{\Gamma}}$, $v \not\in \M{\Tc{\mu(\M{\Gamma})}}$.
\\
Then, we will show:
\\
$(7.0)$\quad $\exists \: \alpha \in \Tc{\mu(\M{\Gamma})}$,
$|\MMM{\Gamma}{\C{\mid\!\sim}{\Gamma}}{\alpha}| < |\mu(\M{\Gamma})|$
\\
But, $\mu(\M{\Gamma}) \subseteq \M{\alpha}$ and,
by $(0)$, $\mu(\M{\Gamma}) \subseteq \MM{\Gamma}{\C{\mid\!\sim}{\Gamma}}$.
Therefore, $\mu(\M{\Gamma}) \subseteq \MMM{\Gamma}{\C{\mid\!\sim}{\Gamma}}{\alpha}$.
\\
Thus, $|\mu(\M{\Gamma})| \leq |\MMM{\Gamma}{\C{\mid\!\sim}{\Gamma}}{\alpha}|$,
which is impossible by $(7.0)$.
\\ \\
{\it Proof of $(7.0)$.} We have $\exists \: \delta \in \Tc{\mu(\M{\Gamma})}$, $v \not\in \M{\delta}$.
\\
By $(A1)$, $|\MMM{\Gamma}{\C{\mid\!\sim}{\Gamma}}{\delta}|$ is finite.
To show $(7.0)$, it suffices to show by induction (in the decreasing direction):
$\forall \: i \in \mathbb{Z}$ with $i \leq |\MMM{\Gamma}{\C{\mid\!\sim}{\Gamma}}{\delta}|$,
\\
$p(i)$\quad $\exists \: \alpha \in \Tc{\mu(\M{\Gamma})}$,
$v \not\in \M{\alpha}$ and
$|\MMM{\Gamma}{\C{\mid\!\sim}{\Gamma}}{\alpha}| - |\mu(\M{\Gamma})| \leq i$.
\\
Obviously, $p(|\MMM{\Gamma}{\C{\mid\!\sim}{\Gamma}}{\delta}|)$ holds (take $\delta$).
\\
Let $i \in \mathbb{Z}$ with $i \leq |\MMM{\Gamma}{\C{\mid\!\sim}{\Gamma}}{\delta}|$
and suppose $p(i)$ holds. We show $p(i-1)$.
\\
We have $\exists \: \alpha \in \Tc{\mu(\M{\Gamma})}$,
$v \not\in \M{\alpha}$ and
$|\MMM{\Gamma}{\C{\mid\!\sim}{\Gamma}}{\alpha}| - |\mu(\M{\Gamma})| \leq i$.
\\
Case~1: $\Td{\MMM{\Gamma}{\C{\mid\!\sim}{\Gamma}}{\alpha}} \subseteq \C{\mid\!\sim}{\Gamma}$.
\\
As $\alpha \in \Tc{\mu(\M{\Gamma})}$ and $(A3)$ holds, we get $\neg\alpha \in \Tc{\mu(\M{\Gamma})}$.
\\
But, $\Tc{\mu(\M{\Gamma})} \cap \Td{\mu(\M{\Gamma})} = \emptyset$. Thus,
neither $\alpha$ nor $\neg\alpha$ belongs to
$\Td{\mu(\M{\Gamma})} = \C{\mid\!\sim}{\Gamma}$.
\\
Consequently, $\alpha \in \G{\Gamma}$. Thus, $v \in \M{\alpha}$, which is impossible.
\\
Case~2: $\exists \: \beta \in \Td{\MMM{\Gamma}{\C{\mid\!\sim}{\Gamma}}{\alpha}}$,
$\beta \not\in \C{\mid\!\sim}{\Gamma}$.
\\
By $(0)$, $\mu(\M{\Gamma}) \subseteq \MM{\Gamma}{\C{\mid\!\sim}{\Gamma}}$.
On the other hand, $\mu(\M{\Gamma}) \subseteq \M{\alpha}$.
Thus, $\mu(\M{\Gamma}) \subseteq \MMM{\Gamma}{\C{\mid\!\sim}{\Gamma}}{\alpha} \subseteq \M{\beta}$.
\\
But, $\beta \not\in \C{\mid\!\sim}{\Gamma} = \Td{\mu(\M{\Gamma})}$. Therefore, $\mu(\M{\Gamma}) \subseteq \M{\neg\beta}$.
\\
Consequently, $\mu(\M{\Gamma}) \subseteq \M{\alpha} \cap \M{\neg\beta} = \M{\alpha \wedge \neg\beta}$
and $\mu(\M{\Gamma}) \subseteq \M{\neg\alpha} \subseteq \M{\neg(\alpha \wedge \neg\beta)}$.
\\
Therefore, $\alpha \wedge \neg\beta \in \Tc{\mu(\M{\Gamma})}$.
\\
Moreover, $v \not\in \M{\alpha} \supseteq \M{\alpha \wedge \neg\beta}$.
\\
In addition,
$\MMM{\Gamma}{\C{\mid\!\sim}{\Gamma}}{\alpha \wedge \neg\beta} \subseteq
\MMM{\Gamma}{\C{\mid\!\sim}{\Gamma}}{\alpha}$, whilst
$\MMM{\Gamma}{\C{\mid\!\sim}{\Gamma}}{\alpha} \not\subseteq
\M{\neg\beta} \supseteq
\MMM{\Gamma}{\C{\mid\!\sim}{\Gamma}}{\alpha \wedge \neg\beta}$.
\\
Thus $|\MMM{\Gamma}{\C{\mid\!\sim}{\Gamma}}{\alpha \wedge \neg\beta}| \leq
|\MMM{\Gamma}{\C{\mid\!\sim}{\Gamma}}{\alpha}| - 1$.
Thus, $|\MMM{\Gamma}{\C{\mid\!\sim}{\Gamma}}{\alpha \wedge \neg\beta}| -
|\mu(\M{\Gamma})| \leq i - 1$.
\\
Therefore, $p(i-1)$ holds (take $\alpha \wedge \neg\beta$).
\\ \\
{\it Proof of $(8)$}. Suppose $\Mp{\Gamma} = \emptyset$.
\\
Now, suppose the contrary of what we want to show, i.e.
$\exists \: v \in \MM{\Gamma}{\C{\mid\!\sim}{\Gamma}}$, $v \not\in \M{\G{\Gamma}}$.
\\
Then, $\exists \: \alpha \in \G{\Gamma}$, $v \not\in \M{\alpha}$.
\\
Case~1: $\alpha \in \T{\MM{\Gamma}{\C{\mid\!\sim}{\Gamma}}}$.
\\
As, $\alpha \in \G{\Gamma}$, $\alpha \not\in \C{\mid\!\sim}{\Gamma}$.
Therefore, by Lemma~\ref{PREF3sim}~$(5)$, $\alpha \not\in \Td{\MM{\Gamma}{\C{\mid\!\sim}{\Gamma}}}$.
\\
Thus, $\alpha \in \Tc{\MM{\Gamma}{\C{\mid\!\sim}{\Gamma}}}$.
Therefore, $\MM{\Gamma}{\C{\mid\!\sim}{\Gamma}} \subseteq \M{\alpha}$.
Consequently, $v \in \M{\alpha}$, which is impossible.
\\
Case~2: $\neg\alpha \in \T{\MM{\Gamma}{\C{\mid\!\sim}{\Gamma}}}$.
\\
As, $\alpha \in \G{\Gamma}$, $\neg\alpha \not\in \C{\mid\!\sim}{\Gamma}$.
Therefore, by Lemma~\ref{PREF3sim}~$(5)$, $\neg\alpha \not\in \Td{\MM{\Gamma}{\C{\mid\!\sim}{\Gamma}}}$.
\\
Thus, $\neg\alpha \in \Tc{\MM{\Gamma}{\C{\mid\!\sim}{\Gamma}}}$.
Therefore, by $(A3)$, $\MM{\Gamma}{\C{\mid\!\sim}{\Gamma}} \subseteq \M{\neg\neg\alpha} = \M{\alpha}$.
\\
Consequently, $v \in \M{\alpha}$, which is impossible.
\\
Case~3: $\alpha \not\in \T{\MM{\Gamma}{\C{\mid\!\sim}{\Gamma}}}$ and
$\neg\alpha \not\in \T{\MM{\Gamma}{\C{\mid\!\sim}{\Gamma}}}$.
\\
Then, by $(A2)$, $\MMM{\Gamma}{\C{\mid\!\sim}{\Gamma}}{\alpha} \not\subseteq \M{\neg\alpha}$.
Thus, $\alpha \in \Td{\MMM{\Gamma}{\C{\mid\!\sim}{\Gamma}}{\alpha}}$.
But, $\alpha \in \G{\Gamma}$. Thus, $\alpha \not\in \C{\mid\!\sim}{\Gamma}$.
\\
Therefore, $\Td{\MMM{\Gamma}{\C{\mid\!\sim}{\Gamma}}{\alpha}} \not\subseteq \C{\mid\!\sim}{\Gamma}$.
Consequently, $\alpha \not\in \G{\Gamma}$, which is impossible.
\\ \\
{\it Proof of $(9)$.}
Case~1: $\Mp{\Gamma} = \emptyset$.
\\
By Lemma~\ref{PREF3sim}~$(6)$,
$\MMM{\Gamma}{\C{\mid\!\sim}{\Gamma}}{\monH{\Gamma}} =
\MMM{\Gamma}{\C{\mid\!\sim}{\Gamma}}{\F{\Gamma}} =
\MM{\Gamma}{\C{\mid\!\sim}{\Gamma}}$.
\\
But, by $(8)$, $(7)$, and $(5)$,
$\MM{\Gamma}{\C{\mid\!\sim}{\Gamma}} =
\MMM{\Gamma}{\C{\mid\!\sim}{\Gamma}}{\G{\Gamma}} =
\MMM{\Gamma}{\C{\mid\!\sim}{\Gamma}}{\Tc{\mu(\M{\Gamma})}} = \mu(\M{\Gamma})$.
\\
Case~2: $\Mp{\Gamma} \not= \emptyset$. Obvious by $(6)$.
\\ \\
{\it Proof of $(10)$.}
\\
Case~1: $\Mp{\Gamma} = \emptyset$.
\\
Case~1.1: $\exists \: v \in \MM{\Gamma}{\C{\mid\!\sim}{\Gamma}}$, $v \not\in \M{\Tc{\mu(\M{\Gamma})}}$.
\\
Case~1.1.1: $\Gamma$ is not consistent.
\\
Then, $\exists \: \alpha \in \Tc{\mu(\M{\Gamma})}$, $v \not\in \M{\alpha}$
and, as $\Gamma$ is not consistent,
$\exists \: \beta \in {\cal F}$, $\M{\Gamma} \subseteq \M{\beta}$ and $\M{\Gamma} \subseteq \M{\neg\beta}$.
\\
We have $\MM{\Gamma}{\C{\mid\!\sim}{\Gamma}} \subseteq \M{\Gamma} \subseteq
\M{\beta} \subseteq \M{\beta \vee \neg\alpha}$.
\\
Moreover, $\mu(\M{\Gamma}) \subseteq \M{\Gamma} \subseteq \M{\neg\beta}$.
Thus, $\mu(\M{\Gamma}) \subseteq \M{\neg\beta} \cap \M{\alpha} = \M{\neg(\beta \vee \neg\alpha)}$.
\\
Therefore, $\beta \vee \neg\alpha \not\in \Td{\mu(\M{\Gamma})} = \C{\mid\!\sim}{\Gamma}$.
\\
In addition, $v \not\in \M{\alpha} \supseteq \M{\neg(\beta \vee \neg\alpha)}$.
\\
Consequently, $v \in \Mi{1}{\Gamma}$ (take $\beta \vee \neg\alpha$ for the $\beta$ of the definition of $\Mi{1}{\Gamma}$).
\\
Thus, $v \in \Mp{\Gamma}$, which is impossible.
\\
Case~1.1.2: $\Gamma$ is consistent.
\\
Thus, $\M{\Gamma} \in {\bf C}$.
Therefore, as $\mu$ is coherency preserving, $\mu(\M{\Gamma}) \in {\bf C}$.
Thus, $\Tc{\mu(\M{\Gamma})} = \emptyset$.
\\
Therefore, $\M{\Tc{\mu(\M{\Gamma})}} = {\cal V}$.
Thus, $v \in \M{\Tc{\mu(\M{\Gamma})}}$, which is impossible.
\\
Case~1.2: $\MM{\Gamma}{\C{\mid\!\sim}{\Gamma}} \subseteq \M{\Tc{\mu(\M{\Gamma})}}$.
\\
Then, by Lemma~\ref{PREF3sim}~$(6)$,
$\MMM{\Gamma}{\C{\mid\!\sim}{\Gamma}}{\monH{\Gamma}} =
\MMM{\Gamma}{\C{\mid\!\sim}{\Gamma}}{\F{\Gamma}} =
\MM{\Gamma}{\C{\mid\!\sim}{\Gamma}} =
\MMM{\Gamma}{\C{\mid\!\sim}{\Gamma}}{\Tc{\mu(\M{\Gamma})}}$.
\\
Therefore, by $(5)$, $\MMM{\Gamma}{\C{\mid\!\sim}{\Gamma}}{\monH{\Gamma}} = \mu(\M{\Gamma})$.
\\
Case~2: $\Mp{\Gamma} \not= \emptyset$. Obvious by $(6)$.\qed
\end{proof}
Now comes the proof of {\bf Proposition~\ref{PREFrepArgSyn}} (which is stated at the beginning of Section~\ref{PREFdpprefdisCR}).

\begin{proof}
{\it Proof of $(0)$.} Direction: ``$\rightarrow$''.
\\
There exists a CP DP coherent choice function $\mu$ from $\bf D$ to ${\cal P}({\cal V})$ such that
\\
$\forall \: \Gamma \subseteq {\cal F}$, $\C{\mid\!\sim}{\Gamma} = \Td{\mu(\M{\Gamma})}$.
\\
We will show:
\\
$(0.0)$\quad $\mid\!\sim$ satisfies $(\mid\!\sim$$0)$.
\\
By Lemma~\ref{PREFPrf2ConArgSyn}~$(1)$, $(2)$, and $(3)$, $\mid\!\sim$ satisfies $(\mid\!\sim$$6)$, $(\mid\!\sim$$7)$, and $(\mid\!\sim$$8)$.
\\
By Lemma~\ref{PREFPrf2ConArgSyn}~$(10)$ and Coherence of $\mu$, $\mid\!\sim$ satisfies $(\mid\!\sim$$9)$.
\\
We will show:
\\
$(0.1)$\quad $\mid\!\sim$ satisfies $(\mid\!\sim$$11)$.

Direction: ``$\leftarrow$''.
\\
Suppose $\mid\!\sim$ satisfies $(\mid\!\sim$$0)$, $(\mid\!\sim$$6)$, $(\mid\!\sim$$7)$, $(\mid\!\sim$$8)$,
$(\mid\!\sim$$9)$, and $(\mid\!\sim$$11)$.
\\
Then, let $\mu$ be the function from ${\bf D}$ to ${\cal P}({\cal V})$ such that
$\forall \: \Gamma \subseteq {\cal F}$, $\mu(\M{\Gamma}) = \MMM{\Gamma}{\C{\mid\!\sim}{\Gamma}}{\monH{\Gamma}}$.
\\
We will show:
\\
$(0.2)$\quad $\mu$ is well-defined.
\\
Clearly, $\mu$ is a DP choice function.
\\
In addition, as $\mid\!\sim$ satisfies $(\mid\!\sim$$9)$, $\mu$ is coherent.
\\
We will show:
\\
$(0.3)$\quad $\mu$ is CP.
\\
And finally, by Lemma~\ref{PREF3sim}~$(7)$, $\forall \: \Gamma \subseteq {\cal F}$,
$\C{\mid\!\sim}{\Gamma} = \Td{\mu(\M{\Gamma})}$.
\\ \\
{\it Proof of $(0.0)$.} Let $\Gamma, \Delta \subseteq {\cal F}$ and suppose
$\C{\vdash}{\Gamma} = \C{\vdash}{\Delta}$.
Then, $\M{\Gamma} = \M{\Delta}$.
\\
Therefore, $\C{\mid\!\sim}{\Gamma} = \Td{\mu(\M{\Gamma})} = \Td{\mu(\M{\Delta})} = \C{\mid\!\sim}{\Delta}$.
\\ \\
{\it Proof of $(0.1)$.} Let $\Gamma \subseteq {\cal F}$ and suppose $\Gamma$ is consistent.
\\
Then, $\M{\Gamma} \in {\bf D} \cap {\bf C}$. Thus, as $\mu$ is CP, $\mu(\M{\Gamma}) \in {\bf C}$.
Therefore, $\Td{\mu(\M{\Gamma})} = \T{\mu(\M{\Gamma})}$.
\\
Consequently, $\Gamma \subseteq \T{\M{\Gamma}} \subseteq \T{\mu(\M{\Gamma})} = \Td{\mu(\M{\Gamma})} = \C{\mid\!\sim}{\Gamma}$.
\\
In addition, $\M{\C{\mid\!\sim}{\Gamma}} = \M{\Td{\mu(\M{\Gamma})}} = \M{\T{\mu(\M{\Gamma})}}$.
But, $\mu(\M{\Gamma}) \in {\bf C}$. Thus, $\M{\T{\mu(\M{\Gamma})}} \in {\bf C}$.
\\
Consequently, $\C{\mid\!\sim}{\Gamma}$ is consistent.
\\
And finally, $\C{\mid\!\sim}{\Gamma} = \Td{\mu(\M{\Gamma})} =
\T{\mu(\M{\Gamma})} = \T{\M{\T{\mu(\M{\Gamma})}}} = \T{\M{\C{\mid\!\sim}{\Gamma}}} = \C{\vdash}{\C{\mid\!\sim}{\Gamma}}$.
\\ \\
{\it Proof of $(0.2)$.} Let $\Gamma, \Delta \subseteq {\cal F}$ and suppose $\M{\Gamma} = \M{\Delta}$.
\\
Then, $\C{\vdash}{\Gamma} = \C{\vdash}{\Delta}$. Thus, by $(\mid\!\sim$$0)$, $\C{\mid\!\sim}{\Gamma} = \C{\mid\!\sim}{\Delta}$.
\\
Consequently, $\monH{\Gamma} = \monH{\Delta}$. Therefore,
$\MMM{\Gamma}{\C{\mid\!\sim}{\Gamma}}{\monH{\Gamma}} = \MMM{\Delta}{\C{\mid\!\sim}{\Delta}}{\monH{\Delta}}$.
\\ \\
{\it Proof of $(0.3)$.} Suppose $V \in {\bf D} \cap {\bf C}$.
Then, $\exists \: \Gamma \subseteq {\cal F}$, $V = \M{\Gamma}$.
\\
Case~1: $\monHi{1}{\Gamma} \not= \emptyset$.
\\
Thus,
$\exists \: \beta \in {\cal F}$, $\beta \not\in \C{\mid\!\sim}{\Gamma}$ and $\MM{\Gamma}{\C{\mid\!\sim}{\Gamma}} \subseteq \M{\beta}$.
\\
By $(\mid\!\sim$$11)$, $\Gamma \subseteq \C{\mid\!\sim}{\Gamma}$ and $\C{\vdash}{\C{\mid\!\sim}{\Gamma}} = \C{\mid\!\sim}{\Gamma}$.
Thus, $\MM{\Gamma}{\C{\mid\!\sim}{\Gamma}} = \M{\C{\mid\!\sim}{\Gamma}}$.
Thus, $\M{\C{\mid\!\sim}{\Gamma}} \subseteq \M{\beta}$.
\\
Therefore, $\beta \in \T{\M{\C{\mid\!\sim}{\Gamma}}} = \C{\vdash}{\C{\mid\!\sim}{\Gamma}} = \C{\mid\!\sim}{\Gamma}$, which is impossible.
\\
Case~2: $\monHi{1}{\Gamma} = \emptyset$.
\\
Then, $\monH{\Gamma} = \emptyset$.
Thus, $\mu(V) = \mu(\M{\Gamma}) = \MMM{\Gamma}{\C{\mid\!\sim}{\Gamma}}{\monH{\Gamma}} =
\M{\C{\mid\!\sim}{\Gamma}}$.
\\
But, by $(\mid\!\sim$$11)$, $\C{\mid\!\sim}{\Gamma}$ is consistent. Therefore,
$\M{\C{\mid\!\sim}{\Gamma}} \in {\bf C}$.
\\ \\ 
{\it proof of $(1)$}. Direction: ``$\rightarrow$''.
\\
Verbatim the proof of $(0)$, except that in addition $\mu$ is LM.
\\
Then, by Lemma~\ref{PREFPrf2ConArgSyn}~$(10)$ and LM, $\mid\!\sim$ satisfies $(\mid\!\sim$$10)$.

Direction: ``$\leftarrow$''.
\\
Verbatim the proof of $(0)$, except that in addition $\mid\!\sim$ satisfies $(\mid\!\sim$$10)$.
\\
Then, by definition of $\mu$ and $(\mid\!\sim$$10)$, $\mu$ is LM.
\\ \\
{\it Proof of $(2)$.} Direction: ``$\rightarrow$''.
\\
Verbatim the proof of $(0)$, except that $\mu$ is no longer CP, whilst $(A2)$ now holds.
\\
Note that, in $(0)$, CP was used only to show $(\mid\!\sim$$11)$ and $(\mid\!\sim$$9)$.
\\
But, $(\mid\!\sim$$11)$ is no longer required to hold.
\\
In addition, by Lemma~\ref{PREFPrf2ConArgSyn}~$(9)$ and Coherence of $\mu$, $(\mid\!\sim$$9)$ holds.

Direction: ``$\leftarrow$''.
\\
Verbatim the proof of $(0)$, except that $(\mid\!\sim$$11)$ does no longer hold, whilst $(A2)$ now holds.
\\
However, in $(0)$, $(\mid\!\sim$$11)$ was used only to show that $\mu$ is CP, which is no longer required.
\\
Note that we do not need to use $(A2)$ in this direction.
\\ \\
{\it Proof of $(3)$.} Direction~``$\rightarrow$''.
\\
Verbatim the proof of $(0)$, except that $\mu$ is no longer CP, whilst $\mu$ is now LM and $(A2)$ now holds.
\\
Note that, in $(0)$, CP was used only to show $(\mid\!\sim$$11)$ and $(\mid\!\sim$$9)$.
\\
But, $(\mid\!\sim$$11)$ is no longer required.
\\
In addition, by Lemma~\ref{PREFPrf2ConArgSyn}~$(9)$ and Coherence of $\mu$, $(\mid\!\sim$$9)$ holds.
\\
Similarly, by Lemma~\ref{PREFPrf2ConArgSyn}~$(9)$ and Local Monotonicity of $\mu$, $(\mid\!\sim$$10)$ holds.

Direction: ``$\leftarrow$''.
\\
Verbatim the proof of $(0)$, except that $(\mid\!\sim$$11)$ does no longer hold,
whilst $(\mid\!\sim$$10)$ and $(A2)$ now holds.
\\
Note that, in $(0)$, $(\mid\!\sim$$11)$ was used only to show that $\mu$ is CP, which is no longer required.
\\
Now, by definition of $\mu$ and by $(\mid\!\sim$$10)$, $\mu$ is LM.
\\
Note that we do not need to use $(A2)$ in this direction.\qed
\end{proof}

\subsection{The discriminative and not necessarily definability preserving case} \label{PREFprefdisCR}

Unlike in Section~\ref{PREFdpprefdisCR}, the conditions of this section will not be purely syntactic.
The translation of properties like Coherence in syntactic terms
is blocked because we do no longer have the following useful equality:
$\mu(\M{\Gamma}) = \MMM{\Gamma}{\C{\mid\!\sim}{\Gamma}}{\monH{\Gamma}}$,
which holds when the choice functions under consideration are definability preserving
(but this is not the case here).
Thanks to Lemmas~\ref{PREFmufestcf} and \ref{PREFprefGen} (stated in Section~\ref{PREFprefCR}), we will provide a solution with semi-syntactic conditions.

\begin{notation}
Let $\cal L$ be a language, $\neg$ a unary connective of $\cal L$, $\cal F$ the set of all wffs of $\cal L$,
$\langle {\cal F}, {\cal V}, \models \rangle$ a semantic structure, and $\mid\!\sim$ a relation on
${\cal P}({\cal F}) \times {\cal F}$.
\\
Then, consider the following condition: $\forall \: \Gamma \subseteq {\cal F}$,
\begin{description}
\item[$(\mid\!\sim$$12)$] $\CCC{\vdash}{\Gamma}{\C{\mid\!\sim}{\Gamma}}{\monH{\Gamma}} =
\T{\lbrace v \in \M{\Gamma} : \forall \: \Delta \subseteq {\cal F},\; \textrm{if}\; v \in \M{\Delta} \subseteq \M{\Gamma},\;\textrm{then}\; v \in \MM{\C{\mid\!\sim}{\Delta}}{\monH{\Delta}} \rbrace}$.
\end{description}
\end{notation}

\begin{proposition} \label{PREFrepGenArg}
Let $\cal L$ be a language, $\neg$ a unary connective of $\cal L$,
$\vee$ and $\wedge$ binary connectives of $\cal L$,
$\cal F$ the set of all wffs of $\cal L$,
$\langle {\cal F}, {\cal V}, \models \rangle$ a semantic structure satisfying $(A1)$ and $(A3)$, and $\mid\!\sim$ a relation on
${\cal P}({\cal F}) \times {\cal F}$. Then,
\begin{description}
\item[$(0)$] $\mid\!\sim$ is a CP preferential-discriminative consequence relation iff
$(\mid\!\sim$$0)$, $(\mid\!\sim$$6)$, $(\mid\!\sim$$7)$, $(\mid\!\sim$$8)$, $(\mid\!\sim$$11)$ and $(\mid\!\sim$$12)$ hold.
\end{description}
Suppose $\langle {\cal F}, {\cal V}, \models \rangle$ satisfies $(A2)$ too. Then,
\begin{description}
\item[$(1)$] $\mid\!\sim$ is a preferential-discriminative consequence relation iff
$(\mid\!\sim$$0)$, $(\mid\!\sim$$6)$, $(\mid\!\sim$$7)$, $(\mid\!\sim$$8)$, and $(\mid\!\sim$$12)$ hold.
\end{description}
\end{proposition}

\begin{proof}

{\it Proof of $(1)$.} Direction: ``$\rightarrow$''.
\\
There exists a coherent choice function $\mu$ from $\bf D$ to ${\cal P}({\cal V})$ such that
$\forall \: \Gamma \subseteq {\cal F}$, $\C{\mid\!\sim}{\Gamma} = \Td{\mu(\M{\Gamma})}$.
\\
Then, $\mid\!\sim$ satisfies obviously $(\mid\!\sim$$0)$.
\\
Let $f$ be the function from ${\bf D}$ to ${\bf D}$ such that
$\forall \: V \in {\bf D}$, $f(V) = \M{\T{\mu(V)}}$.
\\
Then, by Lemma~\ref{PREFprefGen}, $\forall \: V \in {\bf D}$, $f(V) = \M{\T{\mupp{f}{V}}}$.
\\
Moreover, $\forall \: \Gamma \subseteq {\cal F}$,
$f(\M{\Gamma}) = \M{\T{\mu(\M{\Gamma})}} \subseteq \M{\T{\M{\Gamma}}} = \M{\Gamma}$.
\\
Therefore, $f$ is a choice function.
\\
Obviously, $f$ is DP.
\\
In addition, $\forall \: \Gamma \subseteq {\cal F}$,
$\C{\mid\!\sim}{\Gamma} = \Td{\mu(\M{\Gamma})} = \Td{\M{\T{\mu(\M{\Gamma})}}} = \Td{f(\M{\Gamma})}$.
\\
Consequently, by Lemma~\ref{PREFPrf2ConArgSyn}~$(1)$, $(2)$, and $(3)$, $\mid\!\sim$ satisfies
$(\mid\!\sim$$6)$, $(\mid\!\sim$$7)$, and $(\mid\!\sim$$8)$. 
\\
In addition, by Lemma~\ref{PREFPrf2ConArgSyn}~$(9)$, $\forall \: \Gamma \subseteq {\cal F}$, $f(\M{\Gamma}) =
\MMM{\Gamma}{\C{\mid\!\sim}{\Gamma}}{\monH{\Gamma}}$.
\\
We show that $\mid\!\sim$ satisfies $(\mid\!\sim$$12)$. Let $\Gamma \subseteq {\cal F}$.
\\
Then, $\CCC{\vdash}{\Gamma}{\C{\mid\!\sim}{\Gamma}}{\monH{\Gamma}} =
\T{\MMM{\Gamma}{\C{\mid\!\sim}{\Gamma}}{\monH{\Gamma}}} =
\T{f(\M{\Gamma})} = \T{\M{\T{\mupp{f}{\M{\Gamma}}}}} = \T{\mupp{f}{\M{\Gamma}}} =$
\\
$\T{\lbrace v \in \M{\Gamma} : \forall \: W \in {\bf D}$,
if $v \in W \subseteq \M{\Gamma}$, then $v \in f(W) \rbrace} =$
\\
$\T{\lbrace v \in \M{\Gamma} : \forall \: \Delta \subseteq {\cal F}$,
if $v \in \M{\Delta} \subseteq \M{\Gamma}$, then $v \in f(\M{\Delta}) \rbrace} =$
\\
$\T{\lbrace v \in \M{\Gamma} : \forall \: \Delta \subseteq {\cal F}$,
if $v \in \M{\Delta} \subseteq \M{\Gamma}$, then $v \in \MMM{\Delta}{\C{\mid\!\sim}{\Delta}}{\monH{\Delta}} \rbrace} =$
\\
$\T{\lbrace v \in \M{\Gamma} : \forall \: \Delta \subseteq {\cal F},\; \textrm{if}\; v \in \M{\Delta} \subseteq \M{\Gamma},\;\textrm{then}\; v \in \MM{\C{\mid\!\sim}{\Delta}}{\monH{\Delta}} \rbrace}$. 

Direction: ``$\leftarrow$''.
\\
Suppose $(\mid\!\sim$$0)$, $(\mid\!\sim$$6)$, $(\mid\!\sim$$7)$, $(\mid\!\sim$$8)$, and $(\mid\!\sim$$12)$ hold.
\\
Let $f$ be the function from $\bf D$ to $\bf D$ such that
$\forall \: \Gamma \subseteq {\cal F}$, $f(\M{\Gamma}) =
\MMM{\Gamma}{\C{\mid\!\sim}{\Gamma}}{\monH{\Gamma}}$.
\\
By $(\mid\!\sim$$0)$, $f$ is well-defined.
\\
By Lemma~\ref{PREF3sim}~$(7)$, $\forall \: \Gamma \subseteq {\cal F}$, $\C{\mid\!\sim}{\Gamma} = \Td{\MMM{\Gamma}{\C{\mid\!\sim}{\Gamma}}{\monH{\Gamma}}}$.
\\
Therefore, $\forall \: \Gamma \subseteq {\cal F}$, $\C{\mid\!\sim}{\Gamma} = \Td{f(\M{\Gamma})}$.
\\
By $(\mid\!\sim$$12)$, $\forall \: \Gamma \subseteq {\cal F}$, $f(\M{\Gamma}) = \M{\T{\mupp{f}{\M{\Gamma}}}}$.
\\
Therefore, $\forall \: \Gamma \subseteq {\cal F}$, $\C{\mid\!\sim}{\Gamma} = \Td{f(\M{\Gamma})}
= \Td{\M{\T{\mupp{f}{\M{\Gamma}}}}} = \Td{\mupp{f}{\M{\Gamma}}}$.
\\
But, by Lemma~\ref{PREFmufestcf}, $\mup{f}$ is a coherent choice function.
\\ \\
{\it Proof of $(0)$.} Direction: ``$\rightarrow$''.
\\
Verbatim the proof of $(1)$, except that $(A2)$ does no longer hold, whilst $\mu$ is now CP.
\\
Note that $(A2)$ was used only to apply Lemma~\ref{PREFPrf2ConArgSyn}~$(9)$ to get
$\forall \: \Gamma \subseteq {\cal F}$, $f(\M{\Gamma}) =
\MMM{\Gamma}{\C{\mid\!\sim}{\Gamma}}{\monH{\Gamma}}$.
\\
But, we will get this equality by another mean.
\\
Indeed, if $V \in {\bf D} \cap {\bf C}$, then, as $\mu$ is CP, $\mu(V) \in {\bf C}$, thus
$\M{\T{\mu(V)}} \in {\bf C}$, thus $f(V) \in {\bf C}$.
\\
Therefore $f$ is CP.
\\
Consequently, by Lemma~\ref{PREFPrf2ConArgSyn}~$(10)$, we get
$\forall \: \Gamma \subseteq {\cal F}$, $f(\M{\Gamma}) =
\MMM{\Gamma}{\C{\mid\!\sim}{\Gamma}}{\monH{\Gamma}}$.
\\
In addition, by verbatim the proof of $(0.1)$ of Proposition~\ref{PREFrepArgSyn}, $\mid\!\sim$ satisfies $(\mid\!\sim$$11)$.

Direction: ``$\leftarrow$''.
\\
Verbatim the proof of $(1)$, except that $(A2)$ does no longer hold, whilst $\mid\!\sim$ now satisfies
$(\mid\!\sim$$11)$.
\\
But, in this direction, $(A2)$ was not used in $(0)$.
\\
It remains to show that $\mup{f}$ is CP.
\\
By verbatim the proof of $(0.3)$ of Proposition~\ref{PREFrepArgSyn}, we get that $f$ is CP.
\\
Let $V \in {\bf D} \cap {\bf C}$. Then, $f(V) \in {\bf C}$. Thus, $\M{\T{\mupp{f}{V}}} \in {\bf C}$.
Thus, $\mupp{f}{V} \in {\bf C}$ and we are done.\qed
\end{proof}

\section{Conclusion} \label{PREFconclu}

We provided, in a general framework, characterizations for families
of preferential(-discriminative) consequence relations.
Note that we have been strongly inspired by the work of
Schlechta in the non-discriminative case,
whilst we developed new techniques and ideas in the discriminative case. 
In many cases, our conditions are purely syntactic.
In fact, when the choice functions under consideration are not necessarily definability preserving,
we provided solutions with semi-syntactic conditions.
We managed to do so thanks to Lemmas~\ref{PREFmufestcf} and \ref{PREFprefGen}.
An interesting thing is that we used them both in the plain and the discriminative versions.
This suggests that they can be used in yet other versions.
In addition, we are quite confident that Lemmas~\ref{PREF3sim} and \ref{PREFPrf2ConArgSyn}
can be used to characterize other families of consequence relations
defined in the discriminative manner by DP choice functions
(not necessarily coherent, unlike all the families investigated here).

\section{Acknowledgements} \label{PREFack}

I owe very much to Karl Schlechta for his hints, advice, constructive criticism, and more.
I acknowledge also Arnon Avron for valuable discussions.

\bibliography{generalBennaim}
\bibliographystyle{alpha}

\end{document}